\documentclass[10pt,twocolumn,letterpaper]{article}

\usepackage{cvpr}              % To produce the CAMERA-READY version
\usepackage{bm}
\usepackage{mathtools}
\usepackage{pifont}
\usepackage[ruled,vlined]{algorithm2e}
\usepackage{xcolor}
\usepackage[most]{tcolorbox}
\usepackage{multirow}
\usepackage{tcolorbox}

\usepackage{tabularx}
\usepackage{graphicx}
\usepackage{booktabs}
\usepackage{makecell}
\usepackage{arydshln}

\definecolor{cvprblue}{rgb}{0.21,0.49,0.74}
\usepackage[pagebackref,breaklinks,colorlinks,allcolors=cvprblue]{hyperref}

\def\paperID{502} % *** Enter the Paper ID here
\def\confName{CVPR}
\def\confYear{2026}

\title{AnchorFlow: Training-Free 3D Editing via Latent Anchor-Aligned Flows}

\author{
Zhenglin Zhou$^{1}$ \quad
Fan Ma$^{1}$ \quad
Chengzhuo Gui$^{1}$ \quad
Xiaobo Xia$^{2}$ \\
Hehe Fan$^{1}$ \quad
Yi Yang$^{1}$ \quad
Tat-Seng Chua$^{2}$ \\
\\
$^{1}$Zhejiang University \qquad
$^{2}$National University of Singapore
}

\begin{document}

\twocolumn[{
\renewcommand\twocolumn[1][]{#1}%
\maketitle
\vspace{-2.0em}
\begin{center}
    \captionsetup{type=figure}
    \includegraphics[width=1.0\linewidth]{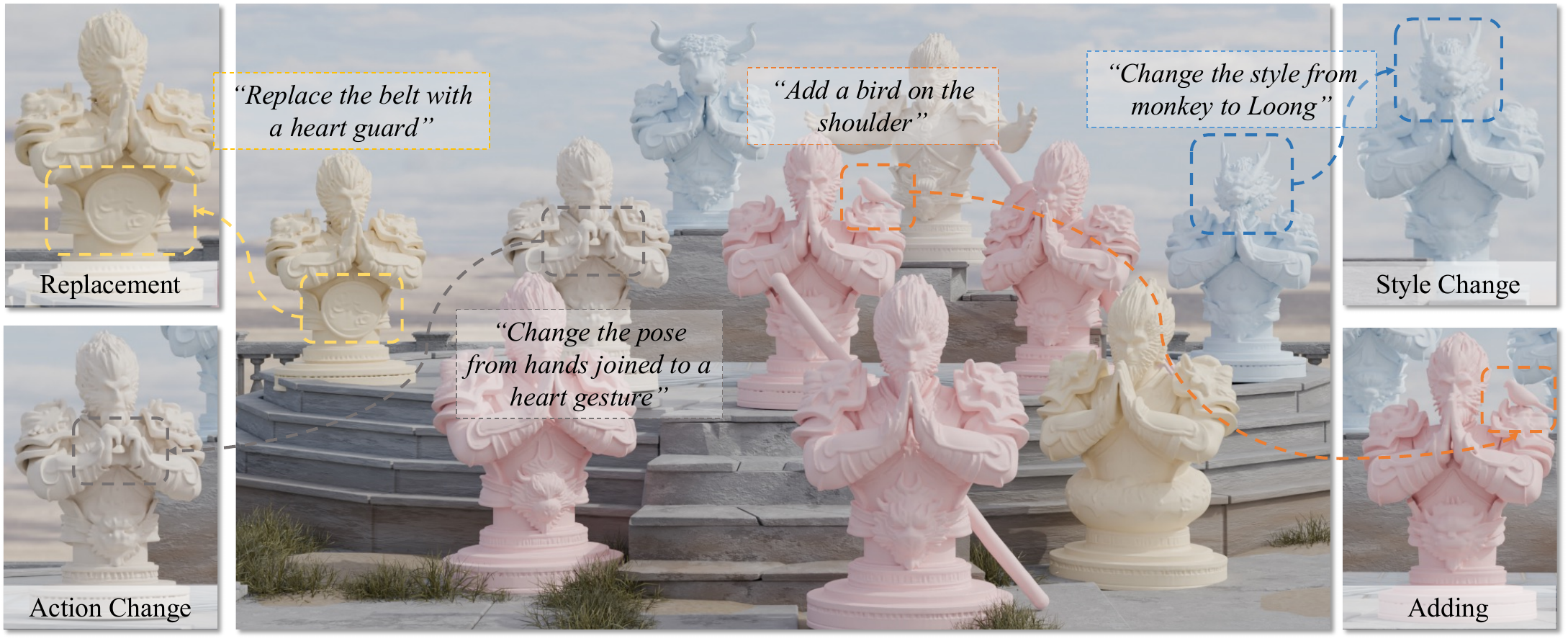}
    % \vspace{-2.0em}
    \captionof{figure}{
\textbf{Overview of 3D Editing Results from AnchorFlow Across Diverse Editing Tasks.}
We present four major types of edits supported by AnchorFlow:
(1) \textbf{Action Change}: altering the pose or articulation of the 3D shape;
(2) \textbf{Object Addition}: introducing new geometric elements;
(3) \textbf{Object Replacement}: substituting existing components with new ones;
(4) \textbf{Style Change}: modifying the shape style while preserving the overall.
    }
    \label{fig:teaser}
\end{center}
    }]

% \maketitle
\begin{abstract}
% Training-free 3D editing aims to modify 3D shapes based on human instructions and plays a crucial role in 3D content creation.
% However, current training-free methods often struggle to produce sufficiently strong or geometrically stable edits due to inconsistent latent anchors arising from timestep-dependent noise.
% To address these limitations, we introduce AnchorFlow, a mask-free and training-free 3D editing framework centered on enforcing latent anchor consistency. 
% AnchorFlow introduces a global anchor shared across source and target trajectories, and employs a relaxed anchor-alignment loss together with an anchor-aligned update rule to ensure coherent transformations throughout the editing process. 
% By stabilizing latent references, AnchorFlow enables stronger semantic modifications and improved geometric fidelity without model finetuning or mask annotations.
% Experiments on the Eval3DEdit benchmark demonstrate that AnchorFlow achieves semantically consistent and structurally stable 3D edits across diverse editing types.
% The code and models will be open-sourced.

Training-free 3D editing aims to modify 3D shapes based on human instructions without model finetuning. 
It plays a crucial role in 3D content creation. 
However, existing approaches often struggle to produce strong or geometrically stable edits, largely due to inconsistent latent anchors introduced by timestep-dependent noise during diffusion sampling.  
To address these limitations, we introduce AnchorFlow, which is built upon the principle of latent anchor consistency. Specifically, AnchorFlow establishes a global latent anchor shared between the source and target trajectories, and enforces coherence using a relaxed anchor-alignment loss together with an anchor-aligned update rule. 
This design ensures that transformations remain stable and semantically faithful throughout the editing process.  
By stabilizing the latent reference space, AnchorFlow enables more pronounced semantic modifications. 
Moreover, AnchorFlow is mask-free. Without mask supervision, it effectively preserves geometric fidelity. 
Experiments on the Eval3DEdit benchmark show that AnchorFlow consistently delivers semantically aligned and structurally robust edits across diverse editing types. Code is at \url{https://github.com/ZhenglinZhou/AnchorFlow}.

\end{abstract}
\section{Introduction}
\label{sec:intro}

3D content editing is essential for applications such as games, films, and AR/VR.
However, producing high-quality edits remains difficult, often requiring extensive manual work and geometric expertise.
This has motivated growing interest in training-free 3D editing, where users specify semantic editing instructions and generative models apply the modification automatically.
Recent progress in 3D flow-based generative models~\cite{xiang2024structured,hunyuan3d2025hunyuan3d} provides strong priors that make such editing feasible.
Yet, despite these advancements, current training-free editing methods often fail to produce sufficiently strong or structurally stable edits, revealing fundamental limitations in existing formulations.

A key challenge lies in how these methods handle inversion within the editing process. 
Current inversion-free strategies~\cite{kulikov2024flowedit} implicitly rely on stochastic Gaussian noise as latent anchors at every timestep. 
Because 3D flow models~\cite{hunyuan3d2025hunyuan3d} are highly sensitive to noise perturbations, these timestep-wise anchors drift unpredictably, leading to inconsistent flow directions.
As a result, semantic changes introduced during editing tend to cancel out, producing insufficient modifications or distorted geometry. 
The reliance on unstable latent anchors restricts the reliability and controllability of training-free 3D editing.

To overcome these limitations, we propose AnchorFlow, a mask-free and training-free 3D editing framework centered on latent anchor consistency. 
Specifically, AnchorFlow introduces a global latent anchor shared by both source and target trajectories, and enforces its consistency through an anchor-alignment loss.
In practice, we optimize a relaxed anchor-alignment loss that encourages the single-step inversions of the two trajectories to remain close in latent space.
Building on this principle, AnchorFlow derives an anchor-aligned update rule that progressively transforms the source shape while preserving its structural identity, without model finetuning or mask annotation.

AnchorFlow can be justified as follows. 
Effective 3D editing demands stable latent references instead of timestep-dependent noise anchors.
By enforcing pairwise consistency at each timestep and leveraging the continuity of the flow, our method naturally propagates these local constraints into a globally consistent latent anchor.
As a result, AnchorFlow enables sufficient semantic modifications with improved geometric fidelity. 
To further evaluate the method, we introduce Eval3DEdit, a benchmark dataset that categorizes 3D edits into five types: addition, removal, replacement, style change, and action change.
Extensive experiments on this benchmark demonstrate that AnchorFlow produces robust, structure-preserving edits and achieves performance comparable to state-of-the-art training-free and mask-free 3D editing methods~\cite{gao2023textdeformer,chen2024generic,bar2025editp23,hunyuan3d2025hunyuan3d,jiao2025uniedit,kulikov2024flowedit}.

In summary, our contributions are as follows:
\begin{itemize}
    \item We introduce an implicit anchor-alignment mechanism that resolves the latent anchor inconsistency problem in inversion-free editing.
    \item We develop AnchorFlow, a mask-free and training-free 3D editing framework that also enables scalable and low-cost curation of pairwise 3D editing data.
    \item We build Eval3DEdit, a benchmark dataset covering representative rigid and non-rigid edits for comprehensive 3D editing evaluation.
\end{itemize}

\begin{figure}[t]
    \centering
    \includegraphics[width=1.0\linewidth]{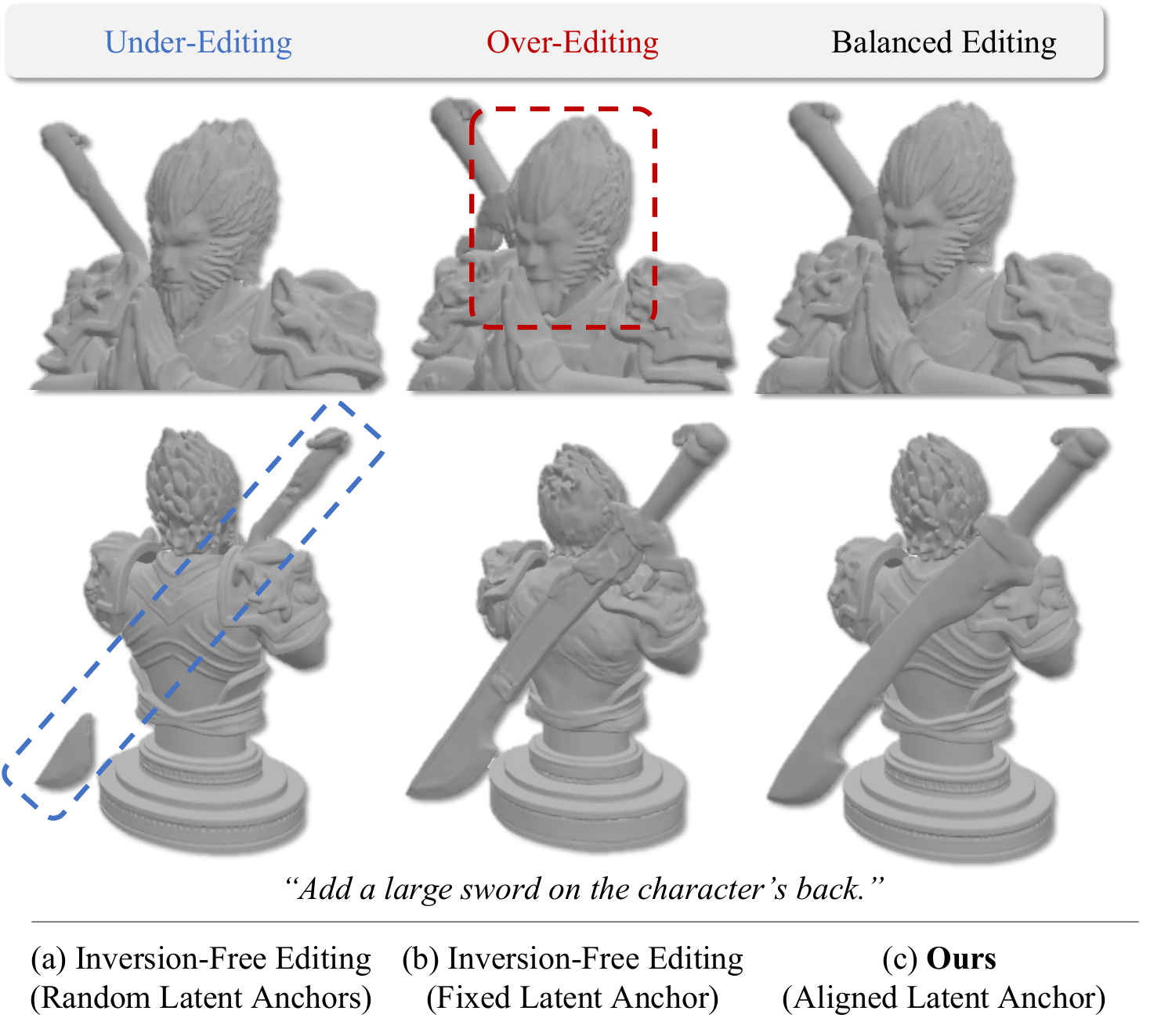}
    \vspace{-2.0em}
    \caption{
    \textbf{Effect of Latent Anchor Selection on 3D Editing.}
    (a) Random timestep-wise anchors cause inconsistent flows, resulting in under-editing and geometric breakage.
    (b) Fixed anchors over-constrain trajectories and push the model away from the source manifold, causing over-editing.
    (c) Our aligned anchors maintain consistent latent references, enabling balanced editing.
    }
    \vspace{-1.0em}
    \label{fig:motivation}
\end{figure}

\section{Related Work}
\label{sec:related_work}

\noindent\textbf{Shape Generation.}
Early progress in 3D generative modeling was constrained by the lack of large-scale 3D data, leading researchers to lift 2D supervision into 3D. Building on text-to-image diffusion priors~\cite{stable_diffusion}, methods such as DreamFields~\cite{jain2021dreamfields}, DreamFusion~\cite{poole2022dreamfusion}, SJC~\cite{wang2022sjc}, and their follow-ups~\cite{wang2023prolificdreamer, yu2023text, zhu2023hifa, katzir2023noise, chung2023luciddreamer, wu2024consistent3d, zhou2025dreamdpo} enabled text-guided 3D generation.
With the emergence of large-scale 3D datasets~\cite{deitke2023objaverse, deitke2024objaverse} and multi-view diffusion models~\cite{liu2023zero, shi2023mvdream, liu2024one, liu2023syncdreamer, long2024wonder3d}, 3D generation became more scalable and efficient. These advances further enabled large reconstruction models (LRMs)~\cite{hong2023lrm, li2023instant3d, zou2024triplane, tang2024lgm, wang2024crm, xu2024grm} that recover 3D assets from multi-view inputs.
Parallel to these developments, native 3D diffusion models have also been explored. 
Some encode 3D shapes into latent spaces using VAEs before diffusion training~\cite{jun2023shap, zhang20233dshape2vecset, zhao2023michelangelo, hong20243dtopia, wu2024direct3d}, while others first obtain neural 3D representations~\cite{zanfir2020neural, kerbl3Dgaussians} and then train diffusion models to generate them~\cite{luo2021diffusion, zhou20213d, nichol2022point, wang2022rodin, jun2023shap, shue20233d, ntavelis2023autodecoding, zhang2024gaussiancube}.
Building on these advancements, large 3D foundation models (LFMs)~\cite{xiang2024structured, zhang2024clay, hunyuan3d22025tencent, hunyuan3d2025hunyuan3d} have recently emerged, showing substantial improvements in quality, efficiency, and robustness. Among them, vecset-based LFMs~\cite{zhang2024clay, hunyuan3d22025tencent, hunyuan3d2025hunyuan3d} stand out for their scalability and practicality, providing strong base capabilities for downstream tasks.

\noindent\textbf{Shape Editing.}
Editing 3D shapes has been a fundamental challenge in computer graphics and vision.
In the early stage, shape editing focuses on geometry-based deformation methods~\cite{sederberg1986free,coquillart1990extended,sorkine2004laplacian}, which enabled smooth local manipulations.
Other classical techniques, including cut-and-paste editing~\cite{biermann2002cut} and sketch-based interfaces~\cite{kanai1999interactive,nealen2005sketch}, allowed interactive mesh fusion and shape manipulation.
While these methods provided geometric control, they required extensive manual intervention, making large-scale edits infeasible.
With the advent of deep learning, the field transitioned toward 2D diffusion-driven 3D editing via score distillation sampling (SDS).
These methods typically harness pretrained 2D diffusion models to guide 3D shape optimization, enabling semantic and text-driven control.
Representative works~\cite{yang2022neumesh,gao2023textdeformer,Ayaan2023instructnerf,sella2023vox,barda2024magicclay,chen2024gaussianeditor,wang2024gaussianeditor,li2024focaldreamer,kim2025meshup} optimize implicit or explicit 3D representations under CLIP or SDS-based guidance.
While these approaches opened a path to generative 3D editing, they suffer from unstable gradients and high computational cost.

More recently, research has moved toward multi-view diffusion and LRMs to overcome the efficiency and consistency limitations of SDS-based optimization.
These methods~\cite{qi2024tailor3d,barda2025instant3dit,erkocc2025preditor3d,li2025cmd,gao20253d,bar2025editp23} generate multi-view images from edited inputs and reconstruct the edited 3D object via a LRM, achieving higher semantic fidelity and efficiency.
However, these pipelines still rely on external diffusion models for view synthesis, which often leads to view-inconsistent artifacts and inaccurate geometry.
Recently, LFMs~\cite{xiang2024structured,hunyuan3d2025hunyuan3d} have achieved significant advances in quality and robustness, offering a new paradigm for 3D creation.
However, their editing capabilities remain underexplored.
Our work takes a step forward by enabling training-free and mask-free 3D shape editing tailored to foundation models.
\section{Method}
\label{sec:method}
In this section, we formally introduce AnchorFlow, a training-free and mask-free 3D editing framework.

\begin{figure*}
    \centering
    \includegraphics[width=1.0\linewidth]{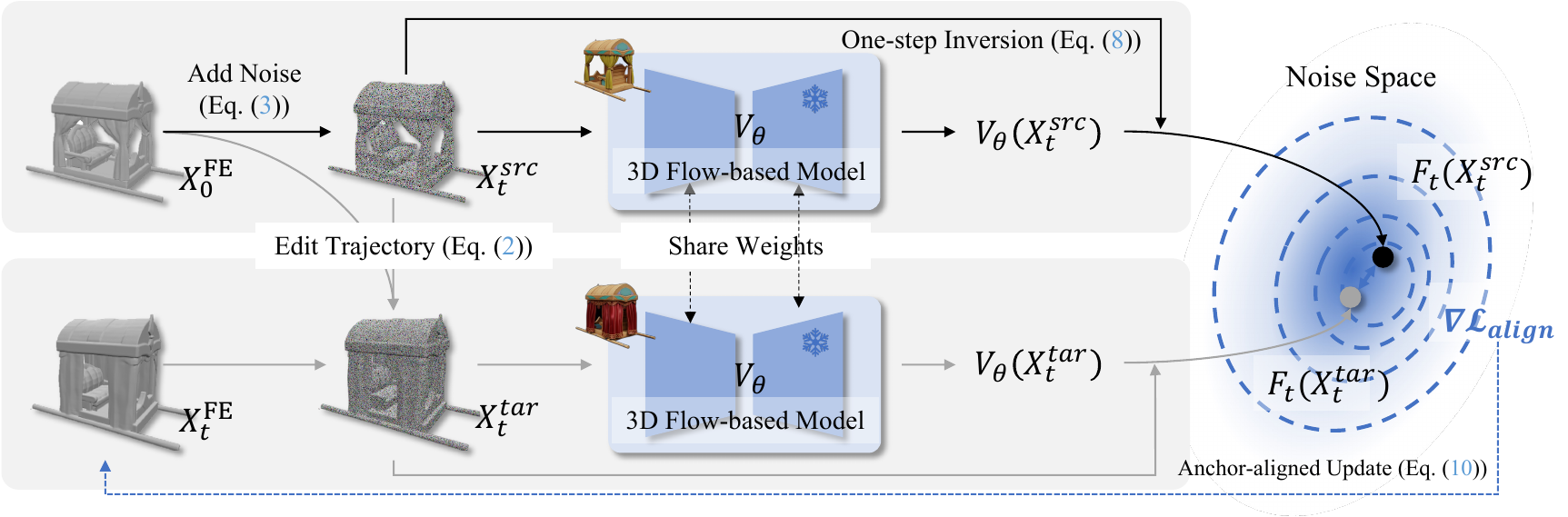}
    \vspace{-2.0em}
    \caption{
    \textbf{Overview of the AnchorFlow for Training-free and Mask-free 3D Editing.}
    Given a source model and an editing instruction, AnchorFlow first constructs the source sample $\bm{X}^{\mathrm{src}}_t$ and forms the editing sample $\bm{X}^{\mathrm{FE}}_t$ at the $t$ step.
    A 3D flow-based model $\bm{v}_\theta$ predicts velocity fields for both the source and target sample.
    To stabilize the editing process, AnchorFlow performs a single-step inversion to approximate the latent anchors $F_t(\bm{X}^{\mathrm{src}}_t)$ and $F_t(\bm{X}^{\mathrm{tar}}_t)$, and aligns them in noise space via the anchor-aligned update guided by $\nabla \mathcal{L}_\mathrm{align}$.
    This design enforces consistent latent anchors, mitigates geometric distortions, and produces structurally stable 3D edits.
    }
    \label{fig:framework}
\end{figure*}

\subsection{Preliminaries}
We first review the formulation of flow matching and its training-free editing extension.

\noindent\textbf{Flow Matching.}
Flow matching formulates generative modeling as learning a continuous-time probability flow from a simple prior distribution to a complex data distribution~\cite{lipman2022flow}.
Let $\{ \bm{X}_t \}_{t \in [ 0, 1]}$ denote the latent trajectory, where $\bm{X}_1$ is commonly initialed from a Gaussian noise and $\bm{X}_0$ corresponds to a data sample (\eg, meshes~\cite{hunyuan3d2025hunyuan3d}).
The dynamics are governed by an ODE:
\begin{equation}
    \mathrm{d} \bm{X}_t = \bm{v}_\theta (\bm{X}_t, t, c) \mathrm{d} t,
\end{equation}
where $\bm{v}_\theta$ is a velocity field conditioned on an optional input $c$.
At inference, samples are obtained by integrating this ODE backward from noise $\bm{X}_1$ to data $\bm{X}_0$.

\noindent\textbf{Inversion-Free Editing.}
Let $\bm{X}^\mathrm{src}$ denote a source asset, and $c_\mathrm{src}$ and $c_\mathrm{tar}$ denote source and target conditions, respectively.
Editing in flow-based generative models aims to obtain an edited asset $\bm{X}^\mathrm{tar}$ that satisfies the semantic modification of $c_\mathrm{tar}$, while faithfully preserving the identity of $\bm{X}^\mathrm{src}$.
FlowEdit~\cite{kulikov2024flowedit} introduces an inversion-free formulation.
It constructs an editing trajectory $\{ \bm{X}^{\mathrm{FE}}_t \}_{t \in [0, 1]}$ that connects the source to the target:
\begin{equation}
    \bm{X}^{\mathrm{FE}}_t = \bm{X}^{\mathrm{tar}}_t - \bm{X}^{\mathrm{src}}_t + \bm{X}^{\mathrm{src}}_0,
\end{equation}
with boundary conditions $\bm{X}^{\mathrm{FE}}_1=\bm{X}^{\mathrm{src}}_0$ and $\bm{X}^{\mathrm{FE}}_0=\bm{X}^{\mathrm{tar}}_0$. 
The editing trajectory $\bm{X}^{\mathrm{FE}}_t$ evolves under the velocity difference $\bm{v}(\bm{X}^{\mathrm{FE}}_t) = \bm{v}_\theta(\bm{X}^{\mathrm{tar}}_t, t, c_\mathrm{tar}) - \bm{v}_\theta(\bm{X}^{\mathrm{src}}_t, t, c_\mathrm{src})$, where the target trajectory is defined as $\bm{X}^{\mathrm{tar}}_t = \bm{X}^{\mathrm{FE}}_t + \bm{X}^{\mathrm{src}}_t - \bm{X}^{\mathrm{src}}_0$ and and the source trajectory is obtained by linearly interpolating between the source latent and Gaussian noise:
\begin{equation}
    \bm{X}^{\mathrm{src}}_t = (1-t)\bm{X}^{\mathrm{src}}_0 + t\bm{N}_t,
    \quad 
    \bm{N}_t \sim \mathcal{N}(0,I).
    \label{eq:add-noise}
\end{equation}
For simplicity, we omit auxiliary arguments in the notation of $\bm{v}(\bm{X}^{\mathrm{FE}}_t)$.
Then, the edited sample is updated using an Euler solver with the velocity difference, where $\delta_t$ denotes the integration step size:
\begin{equation}
    \bm{X}^{\mathrm{FE}}_{t-\delta_t} = \bm{X}^{\mathrm{FE}}_{t} - \delta_t \bm{v}(\bm{X}^{\mathrm{FE}}_t).
\end{equation}

\noindent\textbf{Discussion of Inversion-free Editing.}
While inversion-free editing performs well in 2D image manipulation, directly extending it to 3D foundation models introduces new challenges.
The naive extension often causes insufficient modification (see \cref{fig:motivation} (a)).
A toy experiment helps reveal the underlying cause and motivates our method.

Specifically, we observe that fixing the Gaussian noise during inference alleviates insufficient modification, but alters the object identity (see \cref{fig:motivation} (b)).
This observation suggests that the insufficient modification originates from the random Gaussian noise sampled in the denoising process.
In inversion-free editing, each denoising step samples a new Gaussian noise, resetting the latent anchor at every timestep.
The stochastic flow directions under timestep-wise latent anchors tend to cancel each other out, leading to a near-zero velocity.
Consequently, the update trajectory remains close to the source manifold, resulting in insufficient modification.
Therefore, the key challenge is to stabilize the latent anchor across timesteps, to maintain a consistent latent reference throughout the entire denoising process.
This insight motivates our method, which explicitly aligns source and target trajectories through a global latent anchor, ensuring temporally coherent and geometrically stable edits (see \cref{fig:motivation} (c)).

\subsection{Latent Anchor-Aligned Flows} \label{sec:latent-anchor}

\noindent\textbf{Global Latent Anchor.}
To alleviate the inconsistency across timesteps, we introduce a \emph{global latent anchor} that provides a consistent latent reference for both the source and target trajectories.
We define an ideal latent anchor $\bm{A}$ as a latent point that can simultaneously reconstruct the source and target under their respective conditions:
\begin{equation}
    \bm{A} = F_t(\bm{X}^{\mathrm{src}}_t, t, c_{\mathrm{src}}) 
           = F_t(\bm{X}^{\mathrm{tar}}_t, t, c_{\mathrm{tar}}), \quad \forall\, t\in[0,1],
\end{equation}
where $F_t(\bm{X}_t, t, c)$ denotes the mapping that approximates the latent in the noise space (\eg, the noised latent at $t=1$).
For notational simplicity, we write $F_t(\bm{X}^{\mathrm{src}}_t)$ and $F_t(\bm{X}^{\mathrm{tar}}_t)$ to denote $F_t(\bm{X}^{\mathrm{src}}_t, t, c_{\mathrm{src}})$ and $F_t(\bm{X}^{\mathrm{tar}}_t, t, c_{\mathrm{tar}})$, respectively.
This equality implies that both trajectories share a globally consistent latent anchor throughout the entire flow.

\noindent\textbf{Latent Anchor-based Optimization.}
Directly enforcing $\bm{A}$ to be identical for all timesteps is intractable, since the mapping $F_t$ is implicit in the diffusion dynamics.
We therefore relax this hard constraint into a differentiable least-squares objective by minimizing the deviation of both reconstructions from a global anchor:
\begin{equation}
    \min_{\bm{A}} 
    \sum_{t\in[0,1]} 
    \big(
    \|F_t(\bm{X}^{\mathrm{src}}_t) - \bm{A}\|^2
    + \|F_t(\bm{X}^{\mathrm{tar}}_t) - \bm{A}\|^2
    \big).
\end{equation}
Solving for the optimal $\bm{A}$ yields
$\bm{A}^* = \tfrac{1}{2T}\sum_t [F_t(\bm{X}^{\mathrm{src}}_t)+F_t(\bm{X}^{\mathrm{tar}}_t)]$.
Substituting $\bm{A}^*$ back yields the strong-form latent anchor consistency objective, as detailed in Appendix~\ref{app:anchor-derivation}.
For computational efficiency, we adopt a relaxed form, leading to the following practical latent anchor-alignment loss:
\begin{equation}
    \mathcal{L}_\mathrm{align}
    = \tfrac{1}{2}\sum_{t\in[0,1]}
      \|F_t(\bm{X}^{\mathrm{tar}}_t)
       - F_t(\bm{X}^{\mathrm{src}}_t)\|^2.
    \label{eq:anchor-alignment}
\end{equation}
This relaxation serves as a lower bound of the full objective, enforcing both trajectories to reconstruct toward a global latent reference across all timesteps in an implicit manner.

\noindent\textbf{Single-step Approximation.}
To make $F_t$ tractable, we approximate the inversion using a first-order backward step:
\begin{equation}
    F_t(\bm{X}_t, t, c)
    \approx \bm{X}_t + (1-t)\bm{v}_\theta(\bm{X}_t, t, c),
\end{equation}
which introduces negligible computational overhead while retaining differentiability.
This allows us to express the alignment loss purely in terms of the velocity field.

\noindent\textbf{Latent Anchor-Aligned Update.}
Then, we compute the gradient of $\mathcal{L}_\mathrm{align}$ with respect to the current editing state $\bm{X}^{\mathrm{FE}}_t$:
\begin{equation}
    \begin{aligned}
        \nabla_{\bm{X}^{\mathrm{FE}}_{t}} \mathcal{L}_\mathrm{align} &= \bigg( \frac{\partial F_t(\bm{X}^{\mathrm{tar}}_t)}{\partial \bm{X}^{\mathrm{FE}}_{t}}\bigg)^{\!\top}(F_t(\bm{X}^{\mathrm{tar}}_t) - F_t(\bm{X}^{\mathrm{src}}_t)) \\
    &= \big [ \mathbf{I} + (1-t)\bm{J}_\theta \big ]^{\!\top} (F_t(\bm{X}^{\mathrm{tar}}_t) - F_t(\bm{X}^{\mathrm{src}}_t)),
    \end{aligned}
\end{equation}
where $\bm{J}_\theta = \partial \bm{v}_\theta(\bm{X}^{\mathrm{tar}}_t, t, c_\mathrm{tar})/\partial \bm{X}^{\mathrm{FE}}_t$.
In practice, computing Jacobians for high-dimensional data is expensive. 
Following the Jacobian-free approximation in prior work~\cite{poole2022dreamfusion}, 
we approximate the term $\big[ \mathbf{I} + (1-t)\bm{J}_\theta \big]^\top$ as a scalar multiple of the identity and set
$\big[ \mathbf{I} + (1-t)\bm{J}_\theta \big]^\top \approx (2 - t)\,\mathbf{I}$.
This yields:
\begin{equation}
    \nabla_{\bm{X}^{\mathrm{FE}}_{t}} \mathcal{L}_\mathrm{align} \approx (2-t)(F_t(\bm{X}^{\mathrm{tar}}_t) - F_t(\bm{X}^{\mathrm{src}}_t)).
\end{equation}

% \noindent\textbf{Anchor-aligned Update Formulation.}
Replacing the conventional velocity-difference update with the gradient-derived latent anchor-aligned direction yields:
\begin{equation}
    \bm{X}^{\mathrm{FE}}_{t-\delta_t} = \bm{X}^{\mathrm{FE}}_{t} - \delta_t (2-t) \nabla_{\bm{X}^{\mathrm{FE}}_{t}} \mathcal{L}_\mathrm{align},
\end{equation}
which corresponds to a gradient descent step on the latent anchor alignment loss with respect to the editing state.
This reformulation explicitly aligns the source and target trajectories through a shared latent anchor rather than implicitly relying on stochastic noise correspondence.
Each update step, therefore, becomes geometrically consistent, progressively transforming the source into the target while preserving structural identity.

\subsection{Training-Free 3D Editing}
We detail the implementation of training-free 3D editing with AnchorFlow.

\noindent\textbf{Condition Construction.}
Given a source model and an editing instruction prompt, we first construct the corresponding source condition $c_{\mathrm{src}}$ and target condition $c_{\mathrm{tar}}$.
Specifically, we render 8 image-condition views of the source model following a predefined camera distribution~\cite{xiang2024structured}.
We then employ a large multimodal model (\eg, Gemini-2.5-Flash~\cite{comanici2025gemini}) to rank the alignment between each rendered image and the editing instruction, and select the highest-ranked view as the source condition $c_{\mathrm{src}}$.
Subsequently, an image editing model~\cite{comanici2025gemini} modifies this selected view according to the instruction to generate the target condition $c_{\mathrm{tar}}$.
The conditions $(c_{\mathrm{src}}, c_{\mathrm{tar}})$ are then used to guide the editing process.

\noindent\textbf{AnchorFlow Sampling.}
Given a flow-based 3D generative model $\bm{v}_\theta$~\cite{hunyuan3d2025hunyuan3d}, the source shape (\eg, a mesh) is encoded into its latent code $\bm{X}^{\mathrm{src}}_0$.
With $\bm{X}^{\mathrm{src}}_0$, $(c_{\mathrm{src}}, c_{\mathrm{tar}})$, and $\bm{v}_\theta$, we perform sampling (\cref{alg:proxFM}) to generate an editing trajectory in latent space.
The editing process integrates over $T$ time steps, where editing is from $n_\mathrm{max}$ to $n_\mathrm{min}$, and the balance between source and target guidance is controlled by $s_{\mathrm{src}}$ and $s_{\mathrm{tar}}$.
In our experiments, we use $T = 50$, $s_\mathrm{src} = 3.5$, $s_\mathrm{tar}=7.5$, $n_\mathrm{min} = 1$, and $n_\mathrm{max} = 41$ as default.
The resulting latent $\bm{X}^{\mathrm{FE}}_0$ is finally decoded into 3D space, producing an edited shape that incorporates the desired modification while preserving the source identity.

\begin{algorithm}[t]
\caption{AnchorFlow Sampling}
\label{alg:proxFM}
\KwIn{Source $\bm{X}^{\mathrm{src}}_0$, conditions $c_{\mathrm{src}},c_{\mathrm{tar}}$, flow model $\bm{v}_\theta$, time grid $\{t_i\}_{i=0}^{T}$, schedule $\{\sigma_{t_i}\}_{i=0}^{T}$}
\KwOut{Edited asset $\bm{X}^{\mathrm{FE}}_{0}$}

\BlankLine
\textbf{Initialize:} $\bm{X}^{\mathrm{FE}}_{t_{t_i}}\leftarrow \bm{X}^{\mathrm{src}}_0$\;
\For{$i=T,\dots,1$}{
  $\delta_i \leftarrow \sigma_{t_{i}}-\sigma_{t_i-1}$
  
  $\bm{N}_{t_i}\!\sim\!\mathcal{N}(\mathbf{0},\mathbf{I})$
  
  $\bm{X}^{\mathrm{src}}_{t_i}\!\leftarrow\!(1{-}t_i)\bm{X}^{\mathrm{src}}_0+t_i\bm{N}_{t_i}$
  $\bm{X}^{\mathrm{tar}}_{t_i}\!\leftarrow\!\bm{X}^{\mathrm{FE}}_{t_i}+\bm{X}^{\mathrm{src}}_{t_i}-\bm{X}^{\mathrm{src}}_0$
  
  % $\Delta \bm{V}^{\mathrm{base}}_{t_i}\!\leftarrow\! \bm{v}_\theta(\bm{X}^{\mathrm{tar}}_{t_i},t_i,c_{\mathrm{tar}}) - \bm{v}_\theta(\bm{X}^{\mathrm{src}}_{t_i},t_i,c_{\mathrm{src}})$

  % === NEW LINES START ===
  \tcbox[
    colback=gray!10,
    colframe=black!60,
    title=Anchor-Aligned Update,
    fonttitle=\bfseries,
    boxrule=0.5pt,
    arc=2pt, 
    left=1pt,right=1pt,top=1pt,bottom=1pt
    ]{
    \begin{minipage}{0.95\linewidth}

  $F_{t_i}(\bm{X}^{\mathrm{tar}}_{t_i}) \!\leftarrow\! \bm{X}^{\mathrm{tar}}_{t_i} + (1-t_i) \bm{v}_\theta(\bm{X}^{\mathrm{tar}}_{t_i},t_i,c_{\mathrm{tar}})$
  
  $F_{t_i}(\bm{X}^{\mathrm{src}}_{t_i}) \!\leftarrow\! \bm{X}^{\mathrm{src}}_{t_i} + (1-t_i) \bm{v}_\theta(\bm{X}^{\mathrm{src}}_{t_i},t_i,c_{\mathrm{src}})$

  $\nabla \mathcal{L}_{\mathrm{align}} \leftarrow (2-t_i)(F_{t_i}(\bm{X}^{\mathrm{tar}}_{t_i})-F_{t_i}(\bm{X}^{\mathrm{src}}_{t_i}))$
  
  \end{minipage}
}
  % === NEW LINES END ===

  \textbf{Update:} $\bm{X}^{\mathrm{FE}}_{t_{i-1}}\leftarrow \bm{X}^{\mathrm{FE}}_{t_i} - \delta_i\,\nabla \mathcal{L}_{\mathrm{align}}$\;
}
\Return{$\bm{X}^{\mathrm{FE}}_{0}$}
\end{algorithm}

\section{Experiment}
\label{sec:exp}

\subsection{Experimental Setups}\label{sec:exp_setups}
% In this section, we describe the experimental setups.

\noindent\textbf{Datasets.}
We first construct \textbf{Eval3DEdit}, a benchmark dataset designed for evaluating 3D editing performance.
Eval3DEdit consists of 100 editing samples, evenly distributed across five editing categories: action change, object addition, object removal, object replacement, and style change.
Each sample includes an editing instruction, a source 3D shape, a source image, and a target image.
The source 3D shapes are selected from Objaverse-XL~\cite{deitke2024objaverse} using an aesthetics score threshold of 7.0, ensuring high-quality geometry.
Editing instructions are generated using Gemini 2.5 Pro~\cite{comanici2025gemini} to promote editing diversity and maintain semantic consistency with the source 3D shapes. 
Additional details are provided in the \cref{app:eval3dedit}.

\noindent\textbf{Metrics.}
We quantitatively evaluate the editing performance using two CLIP~\cite{radford2021clip,hessel2021clipscore}-based similarity metrics.
Specifically, $\mathrm{CLIP}_{\text{img}}$ measures the similarity between the rendering from edited 3D shape and the target condition image, reflecting the identity preservation.
$\mathrm{CLIP}_{\text{txt}}$ quantifies the correspondence between the rendering from edited 3D shape and the target editing prompt, reflecting the semantic modification.
% Additional details are provided in the \cref{app:metrics}.
% For each 3D shape, we uniformly render 6 RGB images from different viewpoints, compute the CLIP scores from these multi-view renderings, and average them to obtain the final metric for each shape.

\noindent\textbf{Baselines.}
We compare our approach against seven baseline methods specifically designed for mask-free 3D editing, comprising optimization-based, LRM-based, and LFM-based methods.
Among optimization-based methods, we adopt TextDeformer~\cite{gao2023textdeformer} as the representative approach.
For LRM-based methods, we include MVEdit~\cite{chen2024generic} and EditP23~\cite{bar2025editp23}.
For the LFM-based methods, we adopt Hunyuan3D 2.1~\cite{hunyuan3d2025hunyuan3d} as the base model and implement three representative paradigms:
(1) direct editing, which performs inference directly using the edited image as input;
(2) editing-by-inversion, which first applies flow-based inversion~\cite{jiao2025uniedit} followed by editing inference; and
(3) inversion-free editing, realized through FlowEdit~\cite{kulikov2024flowedit}.
Further implementation details are provided in the \cref{app:baselines}.

\subsection{Comparison with Prior Methods}\label{sec:exp_comparison}

\begin{figure*}[t]
    \centering
    \includegraphics[width=1.0\linewidth]{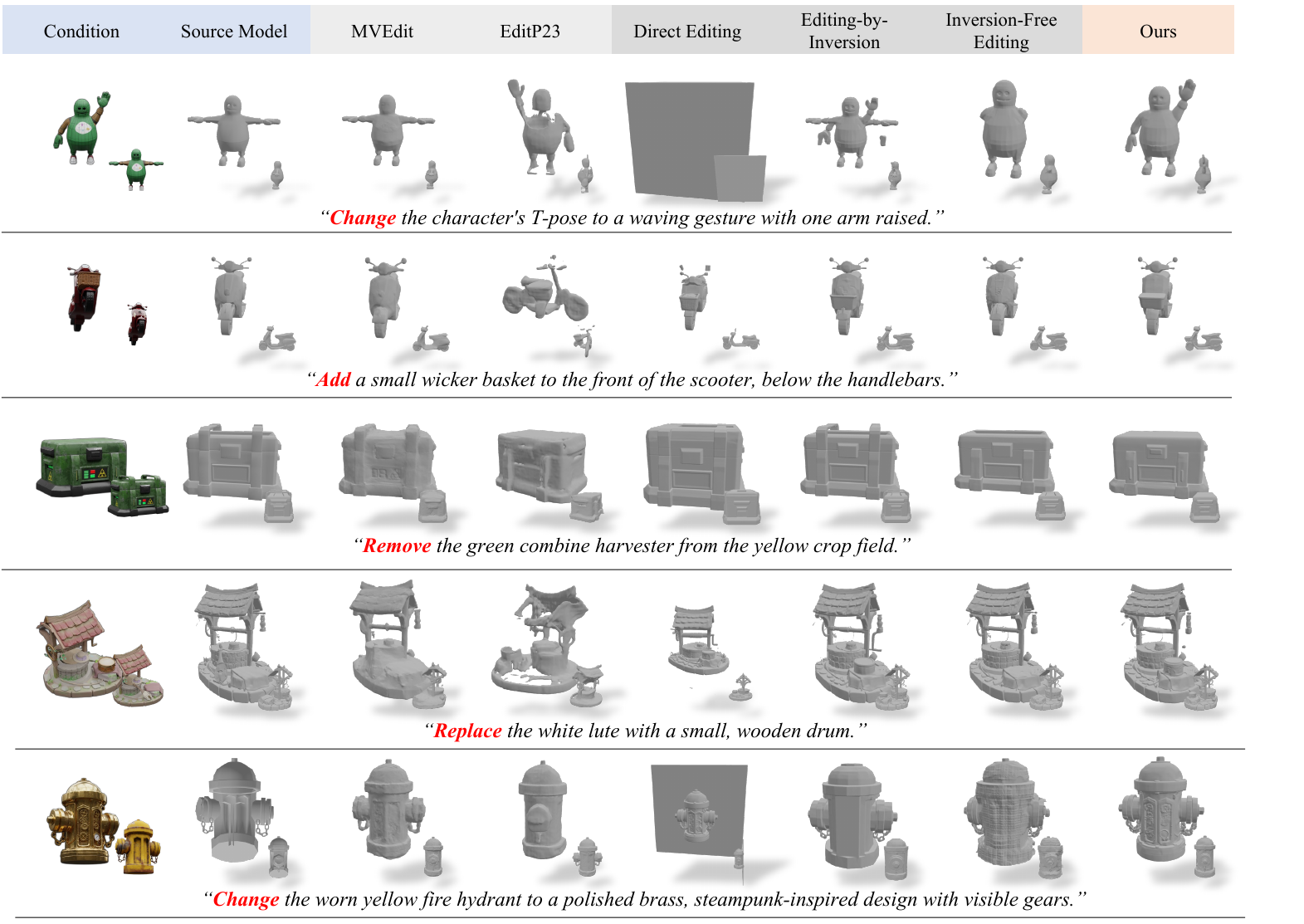}
    \vspace{-1.5em}
    \caption{
    \textbf{Qualitative Comparisons.}
    Each column shows condition pairs, source model, and the corresponding results from various baselines and our method.
    Compared with previous approaches, especially Inversion-Free Editing~\cite{kulikov2024flowedit}, our method produces edits that are both semantically faithful and geometrically consistent, effectively mitigating cases of insufficient edits and distorted geometry.
    % Our method performs well across diverse editing types, including both rigid edits (\eg, addition, removal, replacement) and non-rigid edits (\eg, action or style change).
    }
    \label{fig:comp}
\end{figure*}

\subsubsection{Qualitative Comparisons}
We first evaluate the effectiveness of our method through qualitative comparisons.
As shown in \cref{fig:comp}, our method achieves consistent improvements in both geometric integrity and instruction following.
Compared with the baseline method inversion-free editing~\cite{kulikov2024flowedit}, our method effectively mitigates the issues of insufficient editing (\eg, the second row in \cref{fig:comp}) and geometric distortions in the edited regions (\eg, the first row in \cref{fig:comp}).

Moreover, as a mask-free editing method, we validate the effectiveness of our approach in both rigid and non-rigid editing scenarios.
For rigid editing, including object addition, removal, and replacement, the goal is to enable precise local editing while preserving the unchanged regions.
As shown in the second to fourth rows of \cref{fig:comp}, our method achieves high-quality local edits without explicit masks.
Consequently, our approach facilitates the construction of large-scale pairwise 3D editing datasets at minimal cost, highlighting its strong potential for scalable 3D data curation.
For non-rigid editing, including action and style change, the task requires global transformations while maintaining object identity.
For instance, editing prompts such as ``Change the character's T-pose to a waving gesture with one arm raised.'' (the first row in \cref{fig:comp}) require flexible yet consistent shape adaptation.
Experimental results show that our method exhibits global editing capability while effectively preserving identity, highlighting the advantages of mask-free 3D editing.

\begin{table*}[t]
\centering
\vspace{-0.5em}
\caption{
\textbf{Quantitative Comparison on Eval3DEdit.}
We report results using $\mathrm{CLIP}_{\text{img}} \uparrow$ for identity preservation and $\mathrm{CLIP}_{\text{txt}} \uparrow$ for semantic modification, where higher values indicate better performance.
Metrics are evaluated across five editing categories and the Overall denotes the average performance across all categories.
\textbf{Bold} indicates the best result and \underline{underline} indicates the second-best result.
}
\setlength{\tabcolsep}{2.5pt}
\small
\begin{tabular}{lcccccccccccc}
\toprule
\multirow{2}{*}{Method} & 
\multicolumn{2}{c}{Action Change} & 
\multicolumn{2}{c}{Object Addition} & 
\multicolumn{2}{c}{Object Removal} & 
\multicolumn{2}{c}{Object Replace.} & 
\multicolumn{2}{c}{Style Change} &
\multicolumn{2}{c}{Overall} \\
\cmidrule(lr){2-3} \cmidrule(lr){4-5} \cmidrule(lr){6-7} \cmidrule(lr){8-9} \cmidrule(lr){10-11} \cmidrule(lr){12-13}
& \footnotesize $\mathrm{CLIP}_\text{img}$ & \footnotesize $\mathrm{CLIP}_\text{txt}$ 
& \footnotesize $\mathrm{CLIP}_\text{img}$ & \footnotesize $\mathrm{CLIP}_\text{txt}$ 
& \footnotesize $\mathrm{CLIP}_\text{img}$ & \footnotesize $\mathrm{CLIP}_\text{txt}$ 
& \footnotesize $\mathrm{CLIP}_\text{img}$ & \footnotesize $\mathrm{CLIP}_\text{txt}$ 
& \footnotesize $\mathrm{CLIP}_\text{img}$ & \footnotesize $\mathrm{CLIP}_\text{txt}$ 
& \footnotesize $\mathrm{CLIP}_\text{img}$ & \footnotesize $\mathrm{CLIP}_\text{txt}$ \\
\midrule
\multicolumn{3}{l}{\textcolor{gray}{\textit{Optimization-based 3D Editing Methods}}} & \multicolumn{10}{l}{} \\
TextDeformer~\cite{gao2023textdeformer} & 0.5030 & 0.4389 & 0.5081 & 0.3859 & 0.4969 & 0.4268 & 0.5355 & 0.4281 & 0.4923 & 0.3926 & 0.5074 & 0.4150 \\
\midrule
\multicolumn{3}{l}{\textcolor{gray}{\textit{LRM-based 3D Editing Methods}}} & \multicolumn{10}{l}{} \\
MVEdit~\cite{chen2024generic} & 0.6015 & 0.4346 & 0.4753 & 0.3232 & 0.4784 & 0.3679 & 0.4982 & 0.3511 & 0.4836 & 0.3391 & 0.5074 & 0.3632 \\
EditP23~\cite{bar2025editp23} & 0.5312 & 0.4221 & 0.4561 & 0.3313 & 0.4686 & 0.3918 & 0.4620 & 0.3549 & 0.4697 & 0.3493 & 0.4775 & 0.3699 \\
\midrule
\multicolumn{3}{l}{\textcolor{gray}{\textit{LFM-based 3D Editing Methods}}} & \multicolumn{10}{l}{} \\
Direct Editing~\cite{hunyuan3d2025hunyuan3d} & 0.6709 & \underline{0.4784} & 0.5990 & 0.4287 & 0.5668 & 0.4321 & 0.6737 & 0.4732 & 0.5659 & 0.4131 & 0.6152 & 0.4451 \\
Editing-by-Inversion~\cite{jiao2025uniedit} & \underline{0.7109} & 0.4742 & \textbf{0.7081} & \underline{0.4622} & \underline{0.7282} & \underline{0.4830} & \underline{0.7189} & \underline{0.4825} & \underline{0.6933} & 0.4666 & \underline{0.7119} & \underline{0.4737} \\
Inversion-free Editing~\cite{kulikov2024flowedit} & 0.7096 & 0.4710 & 0.7023 & 0.4441 & \textbf{0.7293} & \textbf{0.4884} & 0.7187 & 0.4787 & 0.6929 & \underline{0.4701} & 0.7106 & 0.4705 \\
\noalign{\vskip 0.25em}
\hdashline
\noalign{\vskip 0.25em}
AnchorFlow (Ours) & \textbf{0.7313} & \textbf{0.5085} & \underline{0.7050} & \textbf{0.4758} & 0.7167 & 0.4735 & \textbf{0.7272} & \textbf{0.4975} & \textbf{0.7061} & \textbf{0.4778} & \textbf{0.7173} & \textbf{0.4866} \\
\bottomrule
\end{tabular}
\vspace{-0.5em}
\label{tab:comp-all}
\end{table*}

\subsubsection{Quantitative Comparisons}
We quantitatively evaluate our method on the Eval3DEdit benchmark.
As shown in \cref{tab:comp-all}, our method achieves the best overall performance, with $\mathrm{CLIP_{img}} = 0.7173$ and $\mathrm{CLIP_{txt}} = 0.4866$.
Compared with the baseline, inversion-free editing~\cite{kulikov2024flowedit}, our method improves the average $\mathrm{CLIP_{img}}$ and $\mathrm{CLIP_{txt}}$ by 0.0067 and 0.0161, respectively, demonstrating a more favorable balance between semantic modification and identity preservation.

Compared with LRM-based 3D editing methods~\cite{chen2024generic,bar2025editp23}, LFM-based methods show improved consistency. This indicates that performing edits through multi-view diffusion models can introduce cross-view inconsistencies, which may reduce editing quality. In contrast, LFM–based methods use end-to-end editing pipelines that operate directly in 3D latent space.
Within the LFM-based 3D editing methods, we conduct three comparative experiments to examine the behavior of different methods.
1) Direct Editing suffers the lowest identity preservation score, indicating that directly generating from the 2D edited image causes the loss of identity information. 
2) Editing-by-Inversion performs well and achieves the second-best results in several categories. Compared with this method, our method maintains comparable performance without using inversion anchors.
3) Inversion-free Editing tends to produce insufficient changes for rigid edits, while introducing geometric distortions in the edited regions for non-rigid edits. The quantitative results show that our method alleviates these issues and achieves balanced performance in both semantic modification and identity preservation.
The result for object removal is moderate, which may be attributed to parameter selection. Adjusting the editing parameters (see \cref{fig:params-quan}) can further improve performance.

\subsection{More Analyses and Justifications}\label{sec:refined_analysis}
% In this section, we conduct further analysis of our method.

\begin{figure}
    \centering
    \includegraphics[width=1.0\linewidth]{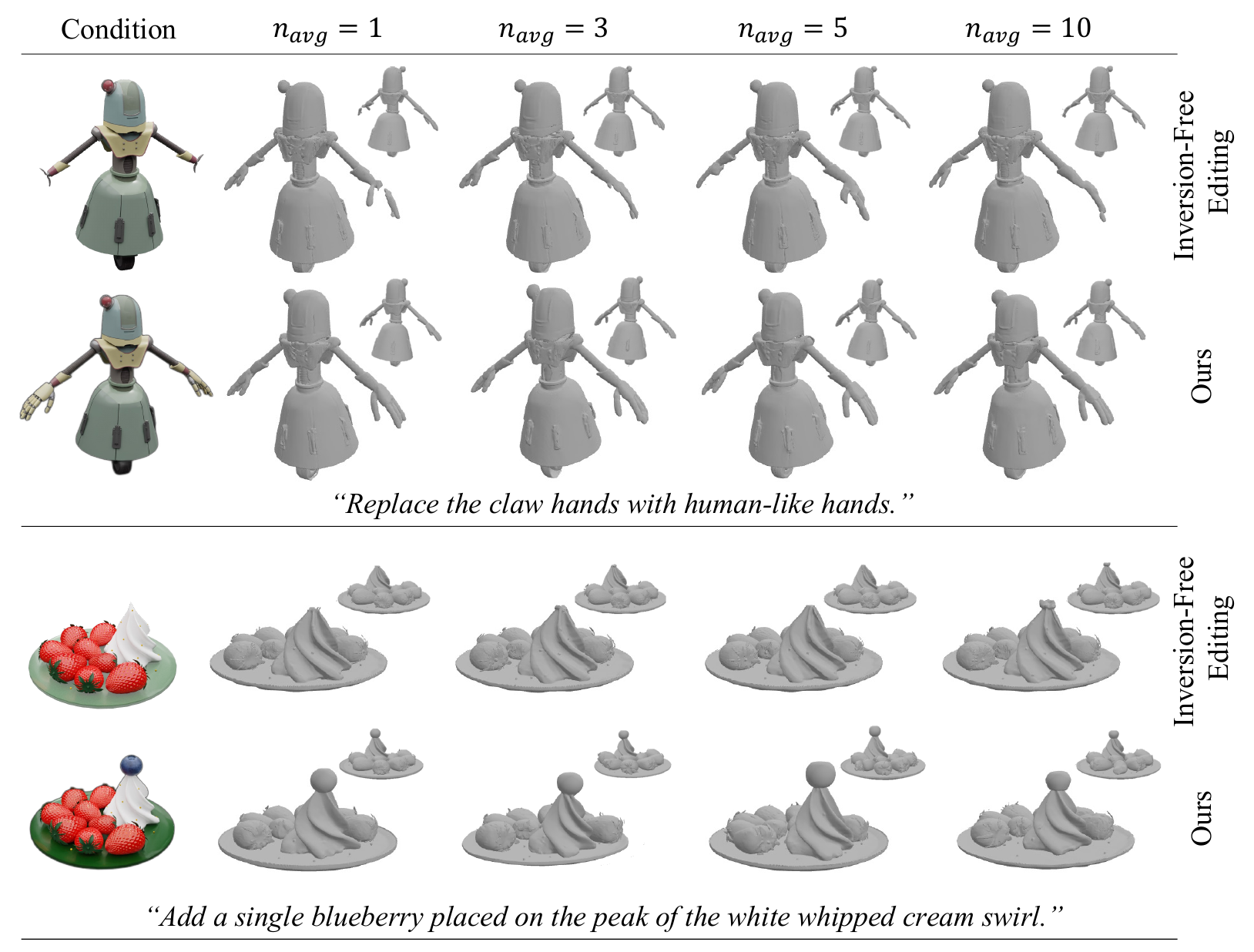}
    % \vspace{-2.0em}
    \caption{
    \textbf{The Effect of Averaging Directions.}
    Averaging $n_\mathrm{avg}$ noisy flow directions stabilizes updates.
    Compared with the computational cost introduced by averaging, AnchorFlow achieves better results with almost no extra time overhead.
    }
    \label{fig:n_avg}
\end{figure}
\noindent\textbf{The Effect of Average Directions.}
Inversion-free editing~\cite{kulikov2024flowedit} proposes to average multiple noisy flow directions to obtain a stable update.
Theoretically, it achieves more faithful edits by averaging the results of multiple random anchors, thereby reducing stochastic deviations and inconsistent directions among them.
This aligns with our motivation to mitigate randomness in anchor estimation, and the two methods are thus compatible.
When combined, they can yield enhanced performance.
\cref{fig:n_avg} shows results for inversion-free editing and our method under different numbers of averaging samples.
The experiments show that both methods benefit from averaging directions.
Moreover, compared with the additional computational cost introduced by averaging, our method achieves a larger performance gain with almost no extra time overhead.

\begin{figure}[t]
    \centering
    \includegraphics[width=1.0\linewidth]{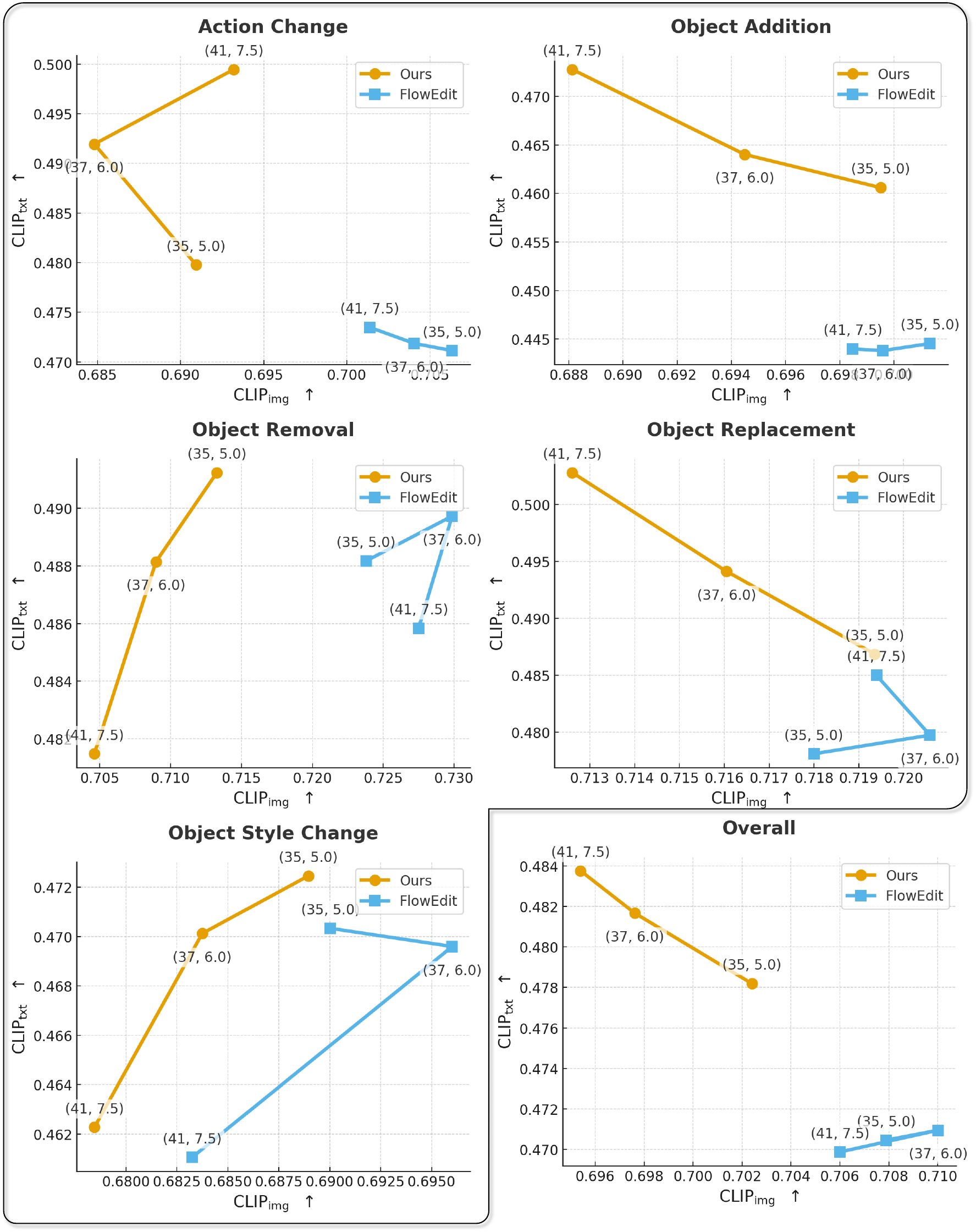}
    \vspace{-1.5em}
    \caption{
    \textbf{Quantitative Analysis of Parameter Selection.}
    We investigate the effect of ($n_\mathrm{max}$, $s_\mathrm{tar}$) across five representative editing categories.
    We report results using $\mathrm{CLIP}_{\text{img}} \uparrow$ for identity preservation and $\mathrm{CLIP}_{\text{txt}} \uparrow$ for semantic modification, where higher values indicate better performance.
    }
    
    \label{fig:params-quan}
\end{figure}

\begin{figure}[t]
    \centering
    \includegraphics[width=1.0\linewidth]{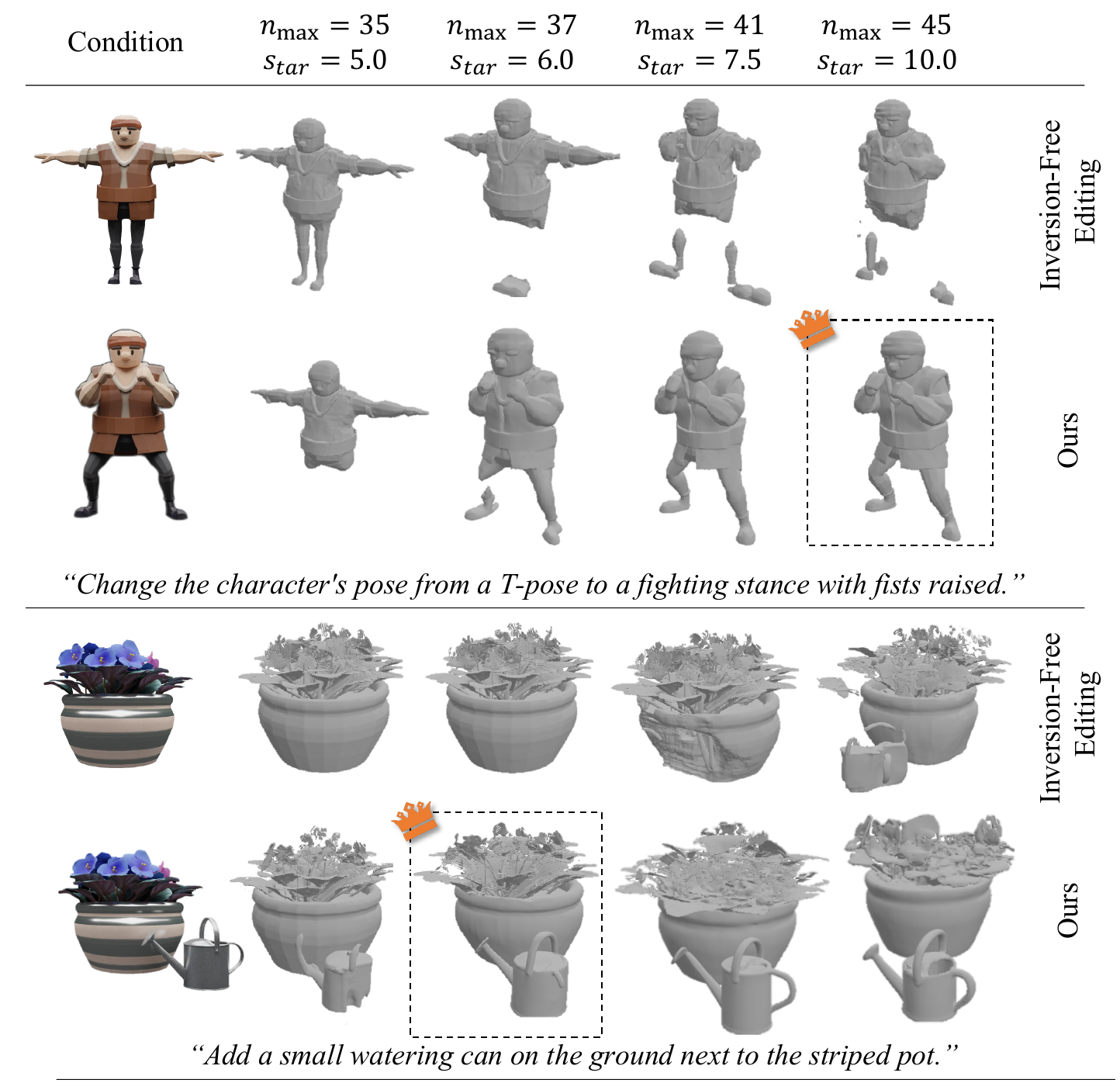}
    \vspace{-1.5em}
    \caption{
    \textbf{Qualitative Analysis of Parameter Selection.}
    We visualize the effect of $(n_\mathrm{max}, s_\mathrm{tar})$.
    Parameter selection varies across editing types, and proper adjustment can improve performance.
    }
    \label{fig:params}
\end{figure}

\noindent\textbf{The Analysis of Parameter Selection.}
We further analyze the influence of editing parameters. 
In our experiments, we fix $T = 50$, $s_\mathrm{src} = 3.5$, and $n_\mathrm{min} = 1$, and investigate how variations in $n_\mathrm{max}$ and $s_\mathrm{tar}$ affect the editing results. 
Specifically, we vary $(n_\mathrm{max}, s_\mathrm{tar})$ among (35, 5.0), (37, 6.0), and (41, 7.5), which jointly control the strength of the editing process..
In general, \cref{fig:params-quan} shows that larger values of $n_\mathrm{max}$ and $s_\mathrm{tar}$ lead to stronger semantic modification, whereas smaller values favor better identity preservation.
A balanced setting of $n_\mathrm{max}=37$ and $s_\mathrm{tar}=6.0$ provides a good trade-off.
Meanwhile, different editing types show distinct parameter preferences, which can be summarized into three patterns.
1) Smaller $(n_\mathrm{max}, s_\mathrm{tar})$. For object removal and object style change, smaller values yield simultaneous improvements in both $\mathrm{CLIP_{img}}$ and $\mathrm{CLIP_{txt}}$.
2) Balanced configuration. For object addition and object replacement, moderate values ($n_\mathrm{max}=37$, $s_\mathrm{tar}=6.0$) achieve balanced results, maintaining structural consistency while introducing intended semantic changes.
3) Larger $(n_\mathrm{max}, s_\mathrm{tar})$. For action change, stronger edits are generally required to accomplish the desired motion transformations, as shown in \cref{fig:params}.
The results demonstrate that parameter selection has a substantial impact on editing performance, and proper adjustment can improve performance.

\noindent\textbf{The Analysis of Time Cost.}
We further analyze the computational efficiency of our method.
As shown in \cref{tab:time}, compared with LRM-based 3D editing methods, LFM-based methods incur lower computational costs due to their one-stage 3D editing process.
Moreover, our method achieves comparable runtime to inversion-free editing~\cite{kulikov2024flowedit}, while achieving higher editing quality.
\begin{table}[h]
\centering
% \vspace{-1.0em}
\caption{
\textbf{Comparison of the Time Cost.}
Our method matches the runtime of \cite{kulikov2024flowedit} while achieving higher editing quality.
}
\vspace{-0.5em}
\resizebox{\columnwidth}{!}{%
\begin{tabular}{lccccccc}
\toprule
& \makecell{Text-\\Deformer~\cite{gao2023textdeformer}} & MVEdit~\cite{chen2024generic} & EditP23~\cite{bar2025editp23} &
  \makecell{Direct\\Editing~\cite{hunyuan3d2025hunyuan3d}} &
  \makecell{Editing-by-\\Inversion~\cite{jiao2025uniedit}} &
  \makecell{Inversion-free\\Editing~\cite{kulikov2024flowedit}} &
  Ours \\
\midrule
Time (s) & 2229.75 & 513.55 & 50.91 & 21.01 & 34.86 & 25.77 & 26.71 \\
\bottomrule
\end{tabular}
}
\vspace{-1.0em}
\label{tab:time}
\end{table}

\subsection{Limitation}
\begin{figure}
    \centering
    \includegraphics[width=1.0\linewidth]{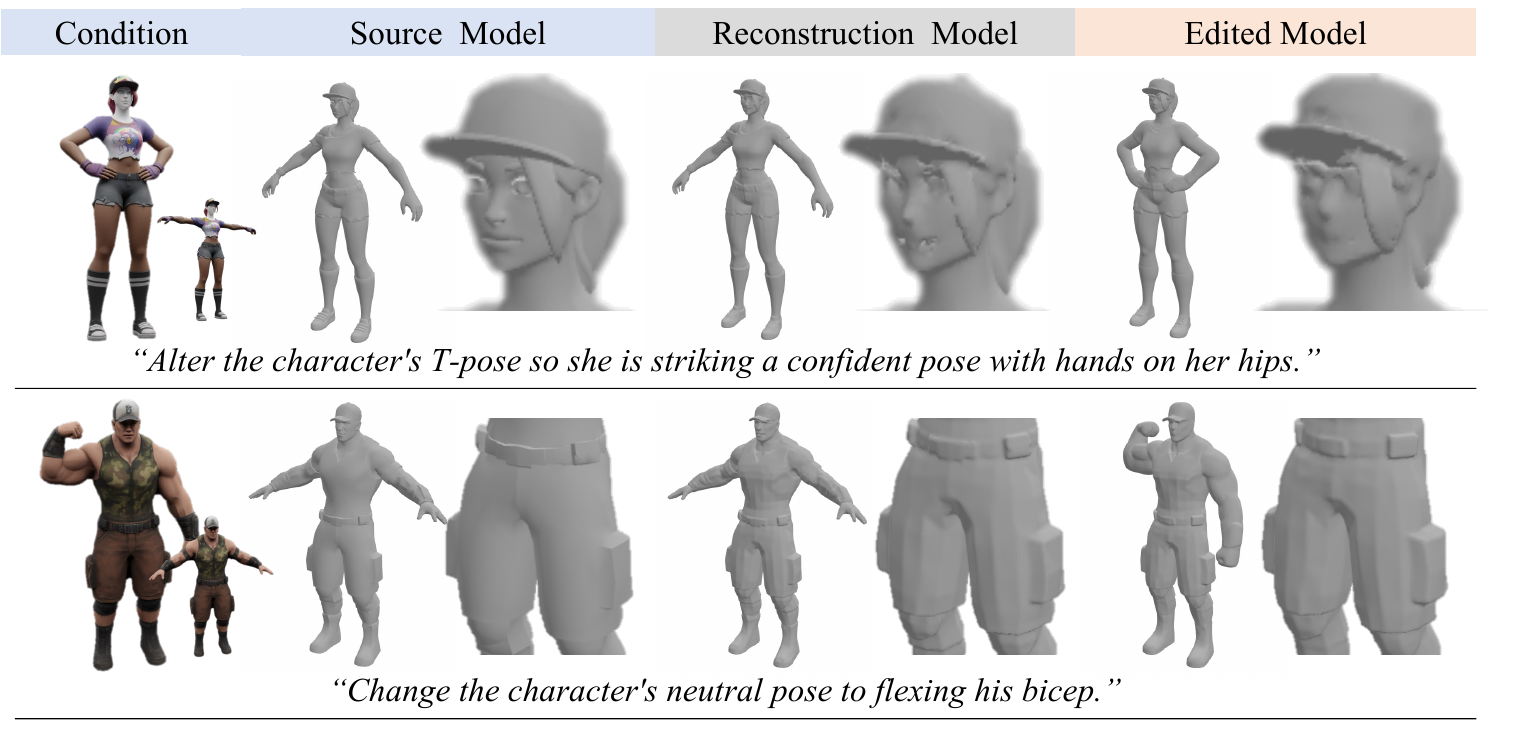}
    \vspace{-1.5em}
    \caption{
    \textbf{Limitation.}
    The reconstruction fidelity of the 3D VAE constrains detail preservation, and the fine features may appear degraded. 
    Future high-fidelity 3D foundation models are expected to alleviate this limitation.
    }
    \label{fig:limitation}
\end{figure}
Although our method shows good editing performance, it is limited by the reconstruction capability of the 3D VAE~\cite{hunyuan3d2025hunyuan3d}. 
Compared with the source model, the reconstructed results lose high-frequency geometry (\eg, facial details in the first row in \cref{fig:limitation}) and structural features (\eg, overalls in the second row in \cref{fig:limitation}). 
As a result, the reconstruction inherently limits detail preservation during editing. 
As 3D foundation models evolve with higher-fidelity latent representations and improved reconstruction quality, our method will benefit from these advances, enabling more accurate and detail-preserving 3D editing.

\section{Conclusion}
\label{sec:conclusion}

In this work, we present AnchorFlow, a training-free and mask-free framework for 3D content editing that resolves the instability caused by timestep-dependent latent anchors. By enforcing global anchor consistency and updating trajectories through anchor-aligned flows, our method enables stronger semantic edits while preserving geometric structure.
Evaluations on the newly constructed Eval3DEdit benchmark demonstrate that AnchorFlow achieves semantically faithful and structurally coherent 3D edits across diverse editing types, establishing a baseline for future explorations in controllable and scalable 3D content creation.

{
    \small
    \bibliographystyle{ieeenat_fullname}
    \bibliography{main}

@String(CVPR= {IEEE Conf. Comput. Vis. Pattern Recog.})

@String(ICCV= {Int. Conf. Comput. Vis.})

@String(ECCV= {Eur. Conf. Comput. Vis.})

@String(TOG= {ACM Trans. Graph.})

@String(ICLR = {Int. Conf. Learn. Represent.})

@String(AAAI = {AAAI})

@article{poole2022dreamfusion,
  title={Dreamfusion: Text-to-3d using 2d diffusion},
  author={Poole, Ben and Jain, Ajay and Barron, Jonathan T and Mildenhall, Ben},
  journal={arXiv preprint arXiv:2209.14988},
  year={2022}
}

@article{li2023instant3d,
  title={Instant3d: Fast text-to-3d with sparse-view generation and large reconstruction model},
  author={Li, Jiahao and Tan, Hao and Zhang, Kai and Xu, Zexiang and Luan, Fujun and Xu, Yinghao and Hong, Yicong and Sunkavalli, Kalyan and Shakhnarovich, Greg and Bi, Sai},
  journal={arXiv preprint arXiv:2311.06214},
  year={2023}
}

@article{shi2023mvdream,
  title={Mvdream: Multi-view diffusion for 3d generation},
  author={Shi, Yichun and Wang, Peng and Ye, Jianglong and Long, Mai and Li, Kejie and Yang, Xiao},
  journal={arXiv preprint arXiv:2308.16512},
  year={2023}
}

@article{nichol2022point,
  title={Point-e: A system for generating 3d point clouds from complex prompts},
  author={Nichol, Alex and Jun, Heewoo and Dhariwal, Prafulla and Mishkin, Pamela and Chen, Mark},
  journal={arXiv preprint arXiv:2212.08751},
  year={2022}
}

@article{jun2023shap,
  title={Shap-e: Generating conditional 3d implicit functions},
  author={Jun, Heewoo and Nichol, Alex},
  journal={arXiv preprint arXiv:2305.02463},
  year={2023}
}

@article{liu2023syncdreamer,
  title={Syncdreamer: Generating multiview-consistent images from a single-view image},
  author={Liu, Yuan and Lin, Cheng and Zeng, Zijiao and Long, Xiaoxiao and Liu, Lingjie and Komura, Taku and Wang, Wenping},
  journal={arXiv preprint arXiv:2309.03453},
  year={2023}
}

@inproceedings{long2024wonder3d,
  title={Wonder3d: Single image to 3d using cross-domain diffusion},
  author={Long, Xiaoxiao and Guo, Yuan-Chen and Lin, Cheng and Liu, Yuan and Dou, Zhiyang and Liu, Lingjie and Ma, Yuexin and Zhang, Song-Hai and Habermann, Marc and Theobalt, Christian and others},
  booktitle={CVPR},
  pages={9970--9980},
  year={2024}
}

@inproceedings{radford2021clip,
  title={Learning transferable visual models from natural language supervision},
  author={Radford, Alec and Kim, Jong Wook and Hallacy, Chris and Ramesh, Aditya and Goh, Gabriel and Agarwal, Sandhini and Sastry, Girish and Askell, Amanda and Mishkin, Pamela and Clark, Jack and others},
  booktitle={ICML},
  pages={8748--8763},
  year={2021},
}

@inproceedings{zhou20213d,
  title     = {3d shape generation and completion through point-voxel diffusion},
  author    = {Zhou, Linqi and Du, Yilun and Wu, Jiajun},
  year      = 2021,
  booktitle = {ICCV},
  pages     = {5826--5835}
}

@article{wang2022rodin,
  title={Rodin: A Generative Model for Sculpting 3D Digital Avatars Using Diffusion},
  author={Wang, Tengfei and Zhang, Bo and Zhang, Ting and Gu, Shuyang and Bao, Jianmin and Baltrusaitis, Tadas and Shen, Jingjing and Chen, Dong and Wen, Fang and Chen, Qifeng and others},
  journal={arXiv preprint arXiv:2212.06135},
  year={2022}
}

@inproceedings{stable_diffusion,
  title={High-resolution image synthesis with latent diffusion models},
  author={Rombach, Robin and Blattmann, Andreas and Lorenz, Dominik and Esser, Patrick and Ommer, Bj{\"o}rn},
  booktitle={CVPR},
  pages={10684--10695},
  year={2022}
}

@article{zanfir2020neural,
	title={Neural Descent for Visual 3D Human Pose and Shape},
	author={Zanfir, Andrei and Bazavan, Eduard Gabriel and Zanfir, Mihai and Freeman, William T and Sukthankar, Rahul and Sminchisescu, Cristian},
	journal={arXiv preprint arXiv:2008.06910},
	year={2020}
}

@article{wang2023prolificdreamer,
  title={ProlificDreamer: High-Fidelity and Diverse Text-to-3D Generation with Variational Score Distillation},
  author={Zhengyi Wang and Cheng Lu and Yikai Wang and Fan Bao and Chongxuan Li and Hang Su and Jun Zhu},
  journal={arXiv preprint arXiv:2305.16213},
  year={2023}
}

@inproceedings{jain2021dreamfields,
  author = {Jain, Ajay and Mildenhall, Ben and Barron, Jonathan T. and Abbeel, Pieter and Poole, Ben},
  title = {Zero-Shot Text-Guided Object Generation with Dream Fields},
  booktitle = {CVPR},
  year = {2022},
}

@inproceedings{wang2022sjc,
      title={Score Jacobian Chaining: Lifting Pretrained 2D Diffusion Models for 3D Generation},
      author={Wang, Haochen and Du, Xiaodan and Li, Jiahao and Yeh, Raymond A. and Shakhnarovich, Greg},
      booktitle={CVPR},
      year={2022},
}

@article{hessel2021clipscore,
  title={Clipscore: A reference-free evaluation metric for image captioning},
  author={Hessel, Jack and Holtzman, Ari and Forbes, Maxwell and Bras, Ronan Le and Choi, Yejin},
  journal={arXiv preprint arXiv:2104.08718},
  year={2021}
}

@Article{kerbl3Dgaussians,
      author       = {Kerbl, Bernhard and Kopanas, Georgios and Leimk{\"u}hler, Thomas and Drettakis, George},
      title        = {3D Gaussian Splatting for Real-Time Radiance Field Rendering},
      journal      = {ACM Transactions on Graphics},
      number       = {4},
      volume       = {42},
      month        = {July},
      year         = {2023}
}

@article{katzir2023noise,
  title={Noise-free score distillation},
  author={Katzir, Oren and Patashnik, Or and Cohen-Or, Daniel and Lischinski, Dani},
  journal={arXiv preprint arXiv:2310.17590},
  year={2023}
}

@inproceedings{chung2023luciddreamer,
  title={Luciddreamer: Domain-free generation of 3d gaussian splatting scenes},
  author={Chung, Jaeyoung and Lee, Suyoung and Nam, Hyeongjin and Lee, Jaerin and Lee, Kyoung Mu},
  booktitle={CVPR},
  year={2024}
}

@article{yu2023text,
  title={Text-to-3d with classifier score distillation},
  author={Yu, Xin and Guo, Yuan-Chen and Li, Yangguang and Liang, Ding and Zhang, Song-Hai and Qi, Xiaojuan},
  journal={arXiv preprint arXiv:2310.19415},
  year={2023}
}

@inproceedings{wu2024consistent3d,
  title={Consistent3d: Towards consistent high-fidelity text-to-3d generation with deterministic sampling prior},
  author={Wu, Zike and Zhou, Pan and Yi, Xuanyu and Yuan, Xiaoding and Zhang, Hanwang},
  booktitle={CVPR},
  year={2024}
}

@article{zhu2023hifa,
  title={Hifa: High-fidelity text-to-3d generation with advanced diffusion guidance},
  author={Zhu, Junzhe and Zhuang, Peiye and Koyejo, Sanmi},
  journal={arXiv preprint arXiv:2305.18766},
  year={2023}
}

@inproceedings{deitke2023objaverse,
  title={Objaverse: A universe of annotated 3d objects},
  author={Deitke, Matt and Schwenk, Dustin and Salvador, Jordi and Weihs, Luca and Michel, Oscar and VanderBilt, Eli and Schmidt, Ludwig and Ehsani, Kiana and Kembhavi, Aniruddha and Farhadi, Ali},
  booktitle={CVPR},
  pages={13142--13153},
  year={2023}
}

@inproceedings{deitke2024objaverse,
  title={Objaverse-xl: A universe of 10m+ 3d objects},
  author={Deitke, Matt and Liu, Ruoshi and Wallingford, Matthew and Ngo, Huong and Michel, Oscar and Kusupati, Aditya and Fan, Alan and Laforte, Christian and Voleti, Vikram and Gadre, Samir Yitzhak and others},
  booktitle={NeurIPS},
  volume={36},
  year={2024}
}

@inproceedings{liu2023zero,
  title={Zero-1-to-3: Zero-shot one image to 3d object},
  author={Liu, Ruoshi and Wu, Rundi and Van Hoorick, Basile and Tokmakov, Pavel and Zakharov, Sergey and Vondrick, Carl},
  booktitle={ICCV},
  pages={9298--9309},
  year={2023}
}

@inproceedings{liu2024one,
  title={One-2-3-45: Any single image to 3d mesh in 45 seconds without per-shape optimization},
  author={Liu, Minghua and Xu, Chao and Jin, Haian and Chen, Linghao and Varma T, Mukund and Xu, Zexiang and Su, Hao},
  booktitle={NeurIPS},
  volume={36},
  year={2024}
}

@article{tang2024lgm,
  title={LGM: Large Multi-View Gaussian Model for High-Resolution 3D Content Creation},
  author={Tang, Jiaxiang and Chen, Zhaoxi and Chen, Xiaokang and Wang, Tengfei and Zeng, Gang and Liu, Ziwei},
  journal={arXiv preprint arXiv:2402.05054},
  year={2024}
}

@inproceedings{zhang2024gaussiancube,
      title={GaussianCube: Structuring Gaussian Splatting using Optimal Transport for 3D Generative Modeling}, 
      author={Bowen Zhang and Yiji Cheng and Jiaolong Yang and Chunyu Wang and Feng Zhao and Yansong Tang and Dong Chen and Baining Guo},
      year={2024},
      booktitle={NeurIPS},
  }

@inproceedings{luo2021diffusion,
  title={Diffusion probabilistic models for 3d point cloud generation},
  author={Luo, Shitong and Hu, Wei},
  booktitle={CVPR},
  pages={2837--2845},
  year={2021}
}

@inproceedings{shue20233d,
  title={3d neural field generation using triplane diffusion},
  author={Shue, J Ryan and Chan, Eric Ryan and Po, Ryan and Ankner, Zachary and Wu, Jiajun and Wetzstein, Gordon},
  booktitle={CVPR},
  pages={20875--20886},
  year={2023}
}

@inproceedings{ntavelis2023autodecoding,
  title={Autodecoding latent 3d diffusion models},
  author={Ntavelis, Evangelos and Siarohin, Aliaksandr and Olszewski, Kyle and Wang, Chaoyang and Gool, Luc V and Tulyakov, Sergey},
  booktitle={NeurIPS},
  pages={67021--67047},
  year={2023}
}

@article{zhang2024clay,
  title={CLAY: A Controllable Large-scale Generative Model for Creating High-quality 3D Assets},
  author={Zhang, Longwen and Wang, Ziyu and Zhang, Qixuan and Qiu, Qiwei and Pang, Anqi and Jiang, Haoran and Yang, Wei and Xu, Lan and Yu, Jingyi},
  journal={ACM Transactions on Graphics},
  volume={43},
  number={4},
  pages={1--20},
  year={2024}
}

@article{hong2023lrm,
  title={Lrm: Large reconstruction model for single image to 3d},
  author={Hong, Yicong and Zhang, Kai and Gu, Jiuxiang and Bi, Sai and Zhou, Yang and Liu, Difan and Liu, Feng and Sunkavalli, Kalyan and Bui, Trung and Tan, Hao},
  journal={arXiv preprint arXiv:2311.04400},
  year={2023}
}

@inproceedings{zou2024triplane,
  title={Triplane meets gaussian splatting: Fast and generalizable single-view 3d reconstruction with transformers},
  author={Zou, Zi-Xin and Yu, Zhipeng and Guo, Yuan-Chen and Li, Yangguang and Liang, Ding and Cao, Yan-Pei and Zhang, Song-Hai},
  booktitle={CVPR},
  pages={10324--10335},
  year={2024}
}

@inproceedings{wang2024crm,
  title={Crm: Single image to 3d textured mesh with convolutional reconstruction model},
  author={Wang, Zhengyi and Wang, Yikai and Chen, Yifei and Xiang, Chendong and Chen, Shuo and Yu, Dajiang and Li, Chongxuan and Su, Hang and Zhu, Jun},
  booktitle={ECCV},
  pages={57--74},
  year={2024},
  organization={Springer}
}

@inproceedings{xu2024grm,
  title={Grm: Large gaussian reconstruction model for efficient 3d reconstruction and generation},
  author={Xu, Yinghao and Shi, Zifan and Yifan, Wang and Chen, Hansheng and Yang, Ceyuan and Peng, Sida and Shen, Yujun and Wetzstein, Gordon},
  booktitle={ECCV},
  pages={1--20},
  year={2024},
  organization={Springer}
}

@article{hong20243dtopia,
  title={3dtopia: Large text-to-3d generation model with hybrid diffusion priors},
  author={Hong, Fangzhou and Tang, Jiaxiang and Cao, Ziang and Shi, Min and Wu, Tong and Chen, Zhaoxi and Yang, Shuai and Wang, Tengfei and Pan, Liang and Lin, Dahua and others},
  journal={arXiv preprint arXiv:2403.02234},
  year={2024}
}

@article{zhang20233dshape2vecset,
  title={3dshape2vecset: A 3d shape representation for neural fields and generative diffusion models},
  author={Zhang, Biao and Tang, Jiapeng and Niessner, Matthias and Wonka, Peter},
  journal={ACM Transactions on Graphics},
  volume={42},
  number={4},
  pages={1--16},
  year={2023},
  publisher={ACM New York, NY, USA}
}

@inproceedings{zhao2023michelangelo,
    title={Michelangelo: Conditional 3D Shape Generation based on Shape-Image-Text Aligned Latent Representation},
    author={Zibo Zhao and Wen Liu and Xin Chen and Xianfang Zeng and Rui Wang and Pei Cheng and BIN FU and Tao Chen and Gang YU and Shenghua Gao},
    booktitle={NeurIPS},
    year={2023},
}

@inproceedings{wu2024direct3d,
  title={Direct3d: Scalable image-to-3d generation via 3d latent diffusion transformer},
  author={Wu, Shuang and Lin, Youtian and Zhang, Feihu and Zeng, Yifei and Xu, Jingxi and Torr, Philip and Cao, Xun and Yao, Yao},
  booktitle={NeurIPS},
  year={2024}
}

@inproceedings{zhou2025dreamdpo,
  title={DreamDPO: Aligning Text-to-3D Generation with Human Preferences via Direct Preference Optimization},
  author={Zhou, Zhenglin and Xia, Xiaobo and Ma, Fan and Fan, Hehe and Yang, Yi and Chua, Tat-Seng},
  booktitle={ICML},
  year={2025}
}

@inproceedings{xiang2024structured,
  title={Structured 3d latents for scalable and versatile 3d generation},
  author={Xiang, Jianfeng and Lv, Zelong and Xu, Sicheng and Deng, Yu and Wang, Ruicheng and Zhang, Bowen and Chen, Dong and Tong, Xin and Yang, Jiaolong},
  booktitle={CVPR},
  year={2025},
}

@inproceedings{zhou2023uni3d,
  title={Uni3d: Exploring unified 3d representation at scale},
  author={Zhou, Junsheng and Wang, Jinsheng and Ma, Baorui and Liu, Yu-Shen and Huang, Tiejun and Wang, Xinlong},
  booktitle={ICLR},
  year={2024}
}

@inproceedings{coquillart1990extended,
  title={Extended free-form deformation: A sculpturing tool for 3D geometric modeling},
  author={Coquillart, Sabine},
  booktitle={SIGGRAPH},
  year={1990}
}

@inproceedings{sederberg1986free,
  title={Free-form deformation of solid geometric models},
  author={Sederberg, Thomas W and Parry, Scott R},
  booktitle={SIGGRAPH},
  year={1986}
}

@article{biermann2002cut,
  title={Cut-and-paste editing of multiresolution surfaces},
  author={Biermann, Henning and Martin, Ioana and Bernardini, Fausto and Zorin, Denis},
  journal={TOG},
  year={2002}
}

@inproceedings{sorkine2004laplacian,
  title={Laplacian surface editing},
  author={Sorkine, Olga and Cohen-Or, Daniel and Lipman, Yaron and Alexa, Marc and R{\"o}ssl, Christian and Seidel, H-P},
  booktitle={SIGGRAPH},
  year={2004}
}

@inproceedings{kanai1999interactive,
  title={Interactive mesh fusion based on local 3D metamorphosis},
  author={Kanai, Takashi and Suzuki, Hiromasa and Mitani, Jun and Kimura, Fumihiko},
  booktitle={Graphics Interface},
  year={1999}
}

@inproceedings{nealen2005sketch,
  title={A sketch-based interface for detail-preserving mesh editing},
  author={Nealen, Andrew and Sorkine, Olga and Alexa, Marc and Cohen-Or, Daniel},
  booktitle={SIGGRAPH},
  year={2005}
}

@inproceedings{yang2022neumesh,
  title={Neumesh: Learning disentangled neural mesh-based implicit field for geometry and texture editing},
  author={Yang, Bangbang and Bao, Chong and Zeng, Junyi and Bao, Hujun and Zhang, Yinda and Cui, Zhaopeng and Zhang, Guofeng},
  booktitle={ECCV},
  year={2022},
}

@inproceedings{sella2023vox,
  title={Vox-e: Text-guided voxel editing of 3d objects},
  author={Sella, Etai and Fiebelman, Gal and Hedman, Peter and Averbuch-Elor, Hadar},
  booktitle={ICCV},
  pages={430--440},
  year={2023}
}

@inproceedings{gao2023textdeformer,
  title={Textdeformer: Geometry manipulation using text guidance},
  author={Gao, William and Aigerman, Noam and Groueix, Thibault and Kim, Vova and Hanocka, Rana},
  booktitle={SIGGRAPH},
  pages={1--11},
  year={2023}
}

@inproceedings{Ayaan2023instructnerf,
     author = {Haque, Ayaan and Tancik, Matthew and Efros, Alexei and Holynski, Aleksander and Kanazawa, Angjoo},
     title = {Instruct-NeRF2NeRF: Editing 3D Scenes with Instructions},
     booktitle = {ICCV},
     year = {2023},
}

@inproceedings{li2024focaldreamer,
  title={Focaldreamer: Text-driven 3d editing via focal-fusion assembly},
  author={Li, Yuhan and Dou, Yishun and Shi, Yue and Lei, Yu and Chen, Xuanhong and Zhang, Yi and Zhou, Peng and Ni, Bingbing},
  booktitle={AAAI},
  year={2024}
}

@inproceedings{wang2024gaussianeditor,
  title={Gaussianeditor: Editing 3d gaussians delicately with text instructions},
  author={Wang, Junjie and Fang, Jiemin and Zhang, Xiaopeng and Xie, Lingxi and Tian, Qi},
  booktitle={CVPR},
  year={2024}
}

@inproceedings{chen2024gaussianeditor,
  title={Gaussianeditor: Swift and controllable 3d editing with gaussian splatting},
  author={Chen, Yiwen and Chen, Zilong and Zhang, Chi and Wang, Feng and Yang, Xiaofeng and Wang, Yikai and Cai, Zhongang and Yang, Lei and Liu, Huaping and Lin, Guosheng},
  booktitle={CVPR},
  year={2024}
}

@inproceedings{barda2024magicclay,
  title={Magicclay: Sculpting meshes with generative neural fields},
  author={Barda, Amir and Kim, Vladimir and Aigerman, Noam and Bermano, Amit Haim and Groueix, Thibault},
  booktitle={SIGGRAPH Asia},
  year={2024}
}

@inproceedings{kim2025meshup,
  title={Meshup: Multi-target mesh deformation via blended score distillation},
  author={Kim, Hyunwoo and Lang, Itai and Aigerman, Noam and Groueix, Thibault and Kim, Vladimir G and Hanocka, Rana},
  booktitle={3DV},
  year={2025}
}

@article{qi2024tailor3d,
  title={Tailor3d: Customized 3d assets editing and generation with dual-side images},
  author={Qi, Zhangyang and Yang, Yunhan and Zhang, Mengchen and Xing, Long and Wu, Xiaoyang and Wu, Tong and Lin, Dahua and Liu, Xihui and Wang, Jiaqi and Zhao, Hengshuang},
  journal={arXiv preprint arXiv:2407.06191},
  year={2024}
}

@article{chen2024generic,
  title={Generic 3d diffusion adapter using controlled multi-view editing},
  author={Chen, Hansheng and Shi, Ruoxi and Liu, Yulin and Shen, Bokui and Gu, Jiayuan and Wetzstein, Gordon and Su, Hao and Guibas, Leonidas},
  journal={arXiv preprint arXiv:2403.12032},
  year={2024}
}

@inproceedings{barda2025instant3dit,
  title={Instant3dit: Multiview inpainting for fast editing of 3d objects},
  author={Barda, Amir and Gadelha, Matheus and Kim, Vladimir G and Aigerman, Noam and Bermano, Amit H and Groueix, Thibault},
  booktitle={CVPR},
  pages={16273--16282},
  year={2025}
}

@inproceedings{erkocc2025preditor3d,
  title={Preditor3d: Fast and precise 3d shape editing},
  author={Erko{\c{c}}, Ziya and G{\"u}meli, Can and Wang, Chaoyang and Nie{\ss}ner, Matthias and Dai, Angela and Wonka, Peter and Lee, Hsin-Ying and Zhuang, Peiye},
  booktitle={CVPR},
  year={2025}
}

@inproceedings{li2025cmd,
  title={Cmd: Controllable multiview diffusion for 3d editing and progressive generation},
  author={Li, Peng and Ma, Suizhi and Chen, Jialiang and Liu, Yuan and Zhang, Congyi and Xue, Wei and Luo, Wenhan and Sheffer, Alla and Wang, Wenping and Guo, Yike},
  booktitle={SIGGRAPH},
  year={2025}
}

@inproceedings{gao20253d,
  title={3d mesh editing using masked lrms},
  author={Gao, William and Wang, Dilin and Fan, Yuchen and Bozic, Aljaz and Stuyck, Tuur and Li, Zhengqin and Dong, Zhao and Ranjan, Rakesh and Sarafianos, Nikolaos},
  booktitle={ICCV},
  pages={7154--7165},
  year={2025}
}

@article{bar2025editp23,
  title={EditP23: 3D Editing via Propagation of Image Prompts to Multi-View},
  author={Bar-On, Roi and Cohen-Bar, Dana and Cohen-Or, Daniel},
  journal={arXiv preprint arXiv:2506.20652},
  year={2025}
}

@misc{hunyuan3d2025hunyuan3d,
    title={Hunyuan3D 2.1: From Images to High-Fidelity 3D Assets with Production-Ready PBR Material},
    author={Tencent Hunyuan3D Team},
    year={2025},
    eprint={2506.15442},
    archivePrefix={arXiv},
    primaryClass={cs.CV}
}

@misc{hunyuan3d22025tencent,
    title={Hunyuan3D 2.0: Scaling Diffusion Models for High Resolution Textured 3D Assets Generation},
    author={Tencent Hunyuan3D Team},
    year={2025},
    eprint={2501.12202},
    archivePrefix={arXiv},
    primaryClass={cs.CV}
}

@article{comanici2025gemini,
  title={Gemini 2.5: Pushing the frontier with advanced reasoning, multimodality, long context, and next generation agentic capabilities},
  author={Comanici, Gheorghe and Bieber, Eric and Schaekermann, Mike and Pasupat, Ice and Sachdeva, Noveen and Dhillon, Inderjit and Blistein, Marcel and Ram, Ori and Zhang, Dan and Rosen, Evan and others},
  journal={arXiv preprint arXiv:2507.06261},
  year={2025}
}

@article{jiao2025uniedit,
    title={UniEdit-Flow: Unleashing Inversion and Editing in the Era of Flow Models}, 
    author={Guanlong Jiao and Biqing Huang and Kuan-Chieh Wang and Renjie Liao},
    journal={arXiv preprint arXiv:2504.13109},
    year={2025},
}

@inproceedings{kulikov2024flowedit,
	title = {FlowEdit: Inversion-Free Text-Based Editing Using Pre-Trained Flow Models},
	author = {Kulikov, Vladimir and Kleiner, Matan and Huberman-Spiegelglas, Inbar and Michaeli, Tomer},
	booktitle = {ICCV},
	year = {2025}
}

@article{lipman2022flow,
  title={Flow matching for generative modeling},
  author={Lipman, Yaron and Chen, Ricky TQ and Ben-Hamu, Heli and Nickel, Maximilian and Le, Matt},
  journal={arXiv preprint arXiv:2210.02747},
  year={2022}
}
}

\clearpage
\onecolumn
\appendix
\section*{Appendix}

The Appendix provides additional technical details, derivations, implementation procedures, and experimental results that complement the main paper. 
It is organized as follows:
\begin{itemize}[leftmargin=2.5em]
    \item \cref{app:anchor-derivation} presents the full derivation of the latent anchor-based optimization framework introduced in the main paper, including the strong-form global objective, its closed-form solution, the reduced objective, and the relaxed per-timestep formulation adopted in practice.
    \item \cref{app:implemtation} provides additional implementation details for our framework, including dataset construction, training-free 3D editing, scalable dataset curation, and the experimental pipeline.
    \item \cref{app:add_settings} summarizes supplementary experimental settings, covering evaluation metrics, baseline implementations, and further comparisons.
    \item \cref{app:results} reports additional qualitative and quantitative results that further validate the effectiveness and robustness of the proposed method.
\end{itemize}

\section{Derivation of the Latent Anchor-based Optimization}
\label{app:anchor-derivation}

In this section, we detail the derivation of the latent anchor-based optimization introduced in \cref{sec:latent-anchor}.

\noindent\textbf{Definition of Latent Anchor-based Optimization.}
As defined in \cref{sec:latent-anchor}, we assume an ideal latent anchor $\bm{A}$.
Let 
\[
\mathbf s_t \triangleq F_t(\bm{X}_t^{\mathrm{src}}, t, c_{\mathrm{src}}),
\qquad
\mathbf g_t \triangleq F_t(\bm{X}_t^{\mathrm{tar}}, t, c_{\mathrm{tar}}),
\]
where $F_t(\cdot)$ maps a latent state at time $t$ to a reference latent (\eg, via a single-step inversion approximation).
The strong form of the global anchor objective is
\begin{equation}
\label{eq:app_strong_obj}
\mathcal J(\bm{A})
= \sum_{t=1}^{T}
\Big(
\|\mathbf s_t - \bm{A}\|_2^2
+ \|\mathbf g_t - \bm{A}\|_2^2
\Big),
\end{equation}
which enforces a single latent anchor $\bm{A}$ to explain both the source and target reconstructions across all timesteps.

Solving for the optimal latent anchor $\bm{A}^*$.
The objective in~\eqref{eq:app_strong_obj} is quadratic and convex in $\bm{A}$.
Setting the gradient to zero gives
\[
\nabla_{\bm{A}}\mathcal J(\bm{A})
= 2\sum_{t=1}^{T}\!\big[(\bm{A}-\mathbf s_t)+(\bm{A}-\mathbf g_t)\big]
= 4T\,\bm{A} - 2\sum_{t=1}^{T}(\mathbf s_t+\mathbf g_t) = \mathbf 0.
\]
Hence,
\begin{equation}
\label{eq:app_Astar}
\boxed{
\bm{A}^* = \frac{1}{2T}\sum_{t=1}^{T}(\mathbf s_t+\mathbf g_t)
}
\end{equation}
That is, the optimal latent anchor is the mean of all per-timestep midpoints $\mathbf m_t \triangleq \tfrac{1}{2}(\mathbf s_t+\mathbf g_t)$.

Substituting $\bm{A}^*$ Back.
To obtain the reduced objective $\mathcal J(\bm{A}^*)$, we use the parallelogram identity:
\begin{equation}
\label{eq:app_parallelogram}
\|\mathbf x -\mathbf m\|^2 + \|\mathbf y-\mathbf m\|^2
= \tfrac{1}{2}\|\mathbf x -\mathbf y\|^2
+ 2\big\|\mathbf m - \tfrac{\mathbf x +\mathbf y}{2}\big\|^2.
\end{equation}
Applying~\eqref{eq:app_parallelogram} with $\mathbf x{=}\mathbf s_t$, $\mathbf y{=}\mathbf g_t$, and $\mathbf m{=}\bm{A}$ yields
\[
\sum_{t=1}^{T}\!\big(\|\mathbf s_t-\bm{A}\|^2+\|\mathbf g_t-\bm{A}\|^2\big)
= \frac{1}{2}\sum_{t=1}^{T}\|\mathbf g_t-\mathbf s_t\|^2
+ 2\sum_{t=1}^{T}\!\Big\|\bm{A} - \frac{\mathbf s_t+\mathbf g_t}{2}\Big\|^2.
\]
Minimizing the right-hand side over $\bm{A}$ gives $\bm{A}^* = \bar{\mathbf m} \triangleq \tfrac{1}{T}\sum_t \mathbf m_t$, and the exact reduced objective is
\begin{equation}
\label{eq:app_reduced_exact}
\boxed{
\mathcal J(\bm{A}^*)
= \frac{1}{2}\sum_{t=1}^{T}\|\mathbf g_t-\mathbf s_t\|^2
+ 2\sum_{t=1}^{T}\|\mathbf m_t - \bar{\mathbf m}\|^2,
\qquad
\mathbf m_t=\tfrac12(\mathbf s_t+\mathbf g_t),\;
\bar{\mathbf m}=\tfrac1T\sum_t \mathbf m_t.
}
\end{equation}
The first term measures pairwise alignment between source and target reconstructions, while the second term enforces the midpoints to remain consistent across timesteps, corresponding to a single global anchor.

\noindent\textbf{Relaxed Latent Anchor-based Optimization.}
For computational efficiency, we adopt a relaxed formulation. Since the second term in \cref{eq:app_reduced_exact} is nonnegative, the objective satisfies
\begin{equation}
\label{eq:app_lowerbound}
\mathcal J(\bm{A}^*) \ge
\frac{1}{2}\sum_{t=1}^{T}\|\mathbf g_t-\mathbf s_t\|^2,
\end{equation}
with equality if and only if $\mathbf m_t = \bar{\mathbf m}$ for all $t$.
In practice, we adopt the simplified per-timestep form
\begin{equation}
\label{eq:app_per_t}
\mathcal L_{\mathrm{align}}
=\tfrac{1}{2}\sum_{t=1}^{T}\|\mathbf g_t-\mathbf s_t\|^2
=\tfrac{1}{2}\sum_{t=1}^{T}\|F_t(\bm{X}_t^{\mathrm{tar}}) - F_t(\bm{X}_t^{\mathrm{src}})\|^2,
\end{equation}
which corresponds to the natural relaxation (\ie, a lower bound) of the strong-form objective~\eqref{eq:app_reduced_exact}, obtained by omitting the temporal midpoint-consistency term.
% Alternatively, using independent per-timestep anchors $\{\bm{A}_t\}$ and minimizing $\sum_t (\|\mathbf s_t-\bm{A}_t\|^2+\|\mathbf g_t-\bm{A}_t\|^2)$ leads exactly to~\eqref{eq:app_per_t}.

% \noindent\textbf{Latent Anchor-based Optimization (Optional)}.
% Define the per-timestep midpoint
% \[
% \hat{\bm{A}}_t \triangleq \tfrac12\big(F_t^{\mathrm{src}}+F_t^{\mathrm{tar}}\big) = \mathbf m_t,
% \quad
% \mathcal L_{\mathrm{temp}}
% =\sum_{t=2}^{T}\|\hat{\bm{A}}_t - \hat{\bm{A}}_{t-1}\|_2^2,
% \]
% and optimize the combined objective
% \[
% \mathcal L_{\mathrm{prox}}
% = \mathcal L_{\mathrm{align}}
% + \lambda_{\mathrm{temp}}\mathcal L_{\mathrm{temp}}.
% \]

\section{Additional Implementation Details}\label{app:implemtation}

\subsection{Details of Eval3DEdit}\label{app:eval3dedit}

Eval3DEdit is a benchmark dataset specifically designed to evaluate 3D editing performance. It comprises 100 samples tailored for 3D editing tasks, categorized into five distinct types: action change, object addition, object removal, object replacement, and style change, with 20 samples allocated to each category. 
Each sample consists of an editing instruction, a source 3D shape, a source image, and a target image.
The dataset is constructed through a multi-stage pipeline that collects high-quality 3D shapes, generates editing instructions, selects optimal source viewpoints, produces target images, and finally curates representative samples for comprehensive 3D editing evaluation.

\begin{itemize}
\item \noindent\textbf{Step 1: Source 3D Shape Collection.}
We collect source models and their corresponding captions from Objaverse-XL~\cite{deitke2024objaverse}. 
To ensure data quality, we filtered for high-quality shapes using an aesthetics score threshold of 7.0.

\item \noindent\textbf{Step 2: Editing Instruction Construction.}
Given the caption associated with each 3D shape, we employed Gemini 2.5 Pro~\cite{comanici2025gemini} to assign an appropriate editing category and to generate a corresponding editing instruction. The prompt used for this process is shown below.
\begin{tcolorbox}[colback=blue!5!white, colframe=blue!75!black, 
    title=Step2: Editing Instruction Construction, width=\textwidth, 
    boxrule=1pt, arc=2mm, auto outer arc,
    sharp corners=south,
    breakable
    ]
\textcolor{blue}{[Task Definition]:} Generation of a Balanced Editing Instruction Dataset

\textcolor{orange}{[Objective]:} 

Given an input CSV file containing assets (\eg, images or videos) and their corresponding textual captions, generate a dataset of editing instructions. This task involves assigning each asset to one of five predefined editing categories and subsequently formulating a precise instruction based on the caption and the assigned category.

\textcolor{olive}{[Editing Categories]:} 

The five (5) mandatory editing categories are:

Object Addition

Object Removal

Object Replacement

Style Change

Action Change

\textcolor{magenta}{[Procedure and Requirements]:}

Data Processing: For each asset in the input CSV, analyze its original caption.

Category Assignment: Assign one of the five predefined editing categories that is semantically appropriate based on the content described in the caption.

Instruction Generation: Based on the original caption and the assigned category, generate a specific, textual "Editing Instruction" detailing the desired modification.

Balanced Distribution: The final output dataset must maintain an approximately equal (balanced) distribution of samples across all five editing categories.

Constraint 1 (Contextual Relevance): All generated instructions (e.g., specifying additions, removals, replacements, or action modifications) must be contextually relevant and semantically coherent with the content of the original asset caption.

Constraint 2 (Specificity): Instructions must be unambiguous and explicit. They must clearly define the specific outcome expected after the edit is applied. Avoid generalized or vague commands. The content to be edited should be specifically designated and clearly defined. For example, for a water cup, the content to be edited should not be a general 'accessory' but rather a specific 'cup handle'.

Constraint 3 (Content Preservation): The edit must not constitute a complete replacement of the original asset's core subject or scene, as this would be equivalent to new generation rather than modification. The core identity of the asset must be preserved.

Constraint 4 (Instructional Diversity): The set of generated instructions must exhibit high diversity. Repetitiveness should be minimized, particularly among instructions generated within the same editing category.

Constraint 5: Action change is only applicable to assets such as humans, animals, and others capable of producing actions. Careful selection of assets suitable for Action change is required when generating editing instructions.

\textcolor{red}{[Output Format]:}

The results should be delivered in a structured CSV file mapping each original asset (and its caption) to its assigned Editing Category and the corresponding generated Editing Instruction. Please generate this CSV file for me.

\end{tcolorbox}

\item \noindent\textbf{Step 3: Source Image Selection.}
We rendered eight viewpoints for each source model using the rendering scripts provided by TRELLIS~\cite{xiang2024structured}.
To select the most suitable source condition for editing, we used Gemini 2.5 Flash to identify the optimal viewpoint.
The prompt used for this selection is provided below.
\begin{tcolorbox}[colback=blue!5!white, colframe=blue!75!black, 
    title=Step3: Source Image Selection, width=\textwidth, 
    boxrule=1pt, arc=2mm, auto outer arc,
    sharp corners=south]
You are an expert visual assistant. You will be given an editing category, a specific instruction, and 8 images of the same object from different angles, labeled Image 0 through Image 7.

Your task is to determine which single image is the most suitable canvas for the requested edit. The best image should provide a clear and direct view of the object or area that needs to be modified.

- Editing Category: "\{category\}"

- Editing Instruction: "\{instruction\}"

Analyze the 8 images provided and identify the best one for this edit.

Your response MUST be a single integer number from 0 to 7, corresponding to the best image. Do not add any other text or explanation.

\end{tcolorbox}

\item \noindent\textbf{Step 4: Target Image Generation.}
Next, we combined the selected source condition with the editing instruction and used Nano Banana~\cite{comanici2025gemini} to generate the target condition. The prompt used for this generation is given below.
\begin{tcolorbox}[colback=blue!5!white, colframe=blue!75!black, 
    title=Step 4: Target Image Generation, width=\textwidth, 
    boxrule=1pt, arc=2mm, auto outer arc,
    sharp corners=south]

As an expert image editor, your task is to edit the following image.

You must strictly and precisely follow the provided category and instruction. The edit should be realistic and seamlessly integrated into the image.

- Editing Category: "\{editing\_category\}"

- Editing Instruction: "\{editing\_instruction\}"

\end{tcolorbox}

\item \noindent\textbf{Step 5: Final Sample Selection.}
To ensure high-quality editing outcomes, we prioritized samples where the source condition represented a frontal viewpoint. Ultimately, we selected 20 samples for each editing category to construct Eval3DEdit, a benchmark dataset covering representative rigid and non-rigid edits for comprehensive 3D editing evaluation.

\end{itemize}

\subsection{Details of Training-Free 3D Editing}\label{app:3d-editing}
This section describes the implementation details of our training-free 3D editing pipeline. 
Given a source model and an editing instruction, we first construct the source and target conditions, and then apply the proposed AnchorFlow sampling procedure to perform 3D editing.

\noindent\textbf{Condition Construction.}
We construct $(c_{\mathrm{src}}, c_{\mathrm{tar}})$ following Step 3 and Step 4 of \cref{app:eval3dedit}. 
We begin by rendering multiple candidate viewpoints of the source model and use an vision language model-based selector~\cite{comanici2025gemini} to determine the optimal source view as $c_{\mathrm{src}}$.
We then pair this selected view with the editing instruction to generate the target condition $c_{\mathrm{tar}}$ using a text-conditioned image editor~\cite{comanici2025gemini}. 
This produces a condition pair that jointly encodes the source appearance and the desired editing semantics.

\noindent\textbf{AnchorFlow Sampling.}
We adopt Hunyuan3D~2.1~\cite{hunyuan3d2025hunyuan3d} as the base shape model $\bm{v}_\theta$. 
Hunyuan3D~2.1 is a high-fidelity 3D generative framework combining a ShapeVAE with a flow-based DiT backbone, and operates on compact vector-set latent representations suitable for 3D geometry encoding.
The source 3D shape is encoded into a latent code $\bm{X}^{\mathrm{src}}_0$. 
Given $\bm{X}^{\mathrm{src}}_0$, the condition pair $(c_{\mathrm{src}}, c_{\mathrm{tar}})$, and the base model $\bm{v}_\theta$, 
we perform AnchorFlow sampling following \cref{alg:proxFM} to generate an editing trajectory in latent space. 
The trajectory integrates over $T$ time steps, evolving from $n_{\mathrm{max}}$ to $n_{\mathrm{min}}$. 
Source and target constraints are balanced by two guidance scales, $s_{\mathrm{src}}$ and $s_{\mathrm{tar}}$, which regulate identity preservation and editing strength, respectively.
We use the default configuration
$T = 50, s_{\mathrm{src}} = 3.5, s_{\mathrm{tar}} = 7.5, n_{\mathrm{min}} = 1, n_{\mathrm{max}} = 41$.
The final latent $\bm{X}^{\mathrm{FE}}_0$ is decoded by Hunyuan3D~2.1 into a mesh, producing an edited shape that incorporates the desired modification while preserving the source identity.
On an NVIDIA H100 (96GB) GPU, each editing instance takes approximately 26.71 s.

\subsection{Scalable 3D Editing Dataset Curation}
Building upon the construction pipeline in \cref{app:eval3dedit} and our training-free 3D editing framework in \cref{app:3d-editing}, we develop a scalable and fully automated procedure for curating large-scale 3D editing datasets. 
Requiring no training or manual mask annotation, the pipeline efficiently produces richly annotated editing pairs.

\noindent\textbf{Condition Pair Construction.}
Our curation pipeline begins by collecting high-quality 3D shapes and their associated captions from 3D shape datasets.
Given the captions, we employ an LLM to assign an editing category and generate a corresponding editing instruction. 
To establish the source condition, we render multiple viewpoints for each 3D shape and use an LLM-based selector to identify the most suitable view. 
The selected view and the editing instruction are then fed into a text-conditioned image editor to generate the corresponding target condition, forming a semantically aligned condition pair.

\noindent\textbf{Training-Free Shape Editing.}
We then leverage our training-free 3D editing framework to synthesize the edited 3D shape. 
The source model is encoded into the latent space of flow-based 3D generative model, and AnchorFlow sampling produces an edited latent trajectory that faithfully applies the specified modification while preserving shape identity. 
The resulting latent is decoded into a mesh, yielding the edited 3D shape paired with its source shape and editing instruction.

\noindent\textbf{Dataset Curation and Applications.}
Our pipeline enables scalable creation of high-quality 3D editing pairs, producing data in the form of (editing instruction, source shape, edited shape).
It offers an efficient approach for constructing large-scale training data for instruction-following 3D editing, thereby supporting the development of more capable and generalizable 3D editing models.

\section{Supplementary Experimental Settings}\label{app:add_settings}

\subsection{Details of Measurement Metrics}\label{app:metrics}
In the main paper, we employ two evaluation strategies to demonstrate the superiority of the proposed method. Here we supplement the details of the measurements.

\subsubsection{Evaluation with CLIP.}
Based on the CLIP (EVA02-E-14-plus) model~\cite{radford2021clip,hessel2021clipscore}, we introduce two CLIP-based similarity metrics $\text{CLIP}_{\text{img}}$ and $\text{CLIP}_{\text{txt}}$ to assess the editing performance on identity preservation and semantic modification, respectively. 

\begin{itemize}
\item \noindent\textbf{Quantitative Evaluation of Identity Preservation.}
Since the image editor~\cite{comanici2025gemini} shows identity-preserving edits, we treat the target image as the ground truth. 
Based on this assumption, we quantitatively evaluate identity preservation by measuring the alignment between the edited 3D shape and the target image. 
Specifically, for each edited shape, we render six views at a resolution of $512 \times 512$, with a $15^\circ$ elevation and a camera radius of $2.4$. 
We then compute the CLIP image similarity between each rendered view and the target image, and use the average similarity score across all views as the identity preservation score $\text{CLIP}_{\text{img}}$.

\item \noindent\textbf{Quantitative Evaluation of Semantic Modification.}
To evaluate semantic modification, we first construct a textual description of the edited shape based on the source shape caption and the editing instruction. 
Specifically, we use the prompt shown below.
We then compute the CLIP image-text similarity between each rendered view and the generated target caption, and use the average similarity across all views as the semantic modification score $\text{CLIP}_{\text{txt}}$.
\begin{tcolorbox}[colback=blue!5!white, colframe=blue!75!black, 
    title=Target Image Caption, width=\textwidth, 
    boxrule=1pt, arc=2mm, auto outer arc,
    sharp corners=south,
    breakable,
    ]

    You are a rewrite assistant. Given the original object captions and an editing instruction, produce ONE English sentence that describes ONLY the final, edited state:
    
    - Describe the final state only, in the format 'subject + change'. Do NOT mention the editing process or use verbs like add/remove/replace/change.
    
    - Be specific and objective; avoid first/second person and irrelevant speculation.
    
    - Output exactly one English sentence. No numbering, no explanations, no quotes.
    
    - Keep it short ($<$= 10 words), lower-case, no quotes, minimal punctuation.
    
    - Examples:
    
    instruction: 'change the robot's pose to be kneeling on one knee.' -$>$ 'robot kneeling on one knee'
    
    instruction: 'add a hat' -$>$ 'figure with hat'
    
    instruction: 'remove the sword' -$>$ 'knight without sword'
    
    instruction: 'replace chair with stool' -$>$ 'scene with stool'

    Editing category: \{category\}

    Editing instruction: \{instruction\}

\end{tcolorbox}
\end{itemize}

\subsubsection{Evaluation with Uni3D} \label{app:metrics-uni3d}
For further validation of identity preservation and semantic modification, we introduce two metrics based on Uni3D~\cite{zhou2023uni3d}: $\text{Uni3D}_{\text{pc}}$ and $\text{Uni3D}_{\text{txt}}$.
Unlike CLIP, Uni3D processes 3D shapes in the form of point clouds. 
All evaluations are conducted using the Uni3D (eva\_giant\_patch14\_560) model.

\begin{itemize}
\item \noindent\textbf{Quantitative Evaluation of Identity Preservation.}
We measure identity preservation using the Uni3D similarity between the source shape and the edited shape. 
Specifically, we extract point clouds from both shapes and assign them a fixed color (\eg, gray with RGB values (100, 100, 100)). 
Features are then extracted using Uni3D, and identity preservation is computed as the cosine similarity between the two point-cloud feature vectors.

\item \noindent\textbf{Quantitative Evaluation of Semantic Modification.}
Similar to $\text{CLIP}_{\text{txt}}$, we evaluate semantic modification $\text{Uni3D}_{\text{txt}}$ by computing the similarity between the Uni3D point-cloud feature of the edited shape and the target caption.
\end{itemize}

\subsection{Details of Baselines}\label{app:baselines}
In this section, we give the implementation details of the baseline methods, including TextDeformer~\cite{gao2023textdeformer}, MVEdit~\cite{chen2024generic}, EditP23~\cite{bar2025editp23}, Direct Editing~\cite{hunyuan3d2025hunyuan3d}, Editing-by-Inversion~\cite{jiao2025uniedit}, and Inversion-free Editing~\cite{kulikov2024flowedit}.  

\begin{itemize}
\item \noindent\textbf{Implementation Details of TextDeformer.}
We reproduce TextDeformer~\cite{gao2023textdeformer} following the official implementation\footnote{\url{https://github.com/threedle/TextDeformer}}. 
TextDeformer deforms a source mesh into a text-specified target shape using differentiable rendering and CLIP-based semantic alignment. 
Instead of directly optimizing vertex displacements, it optimizes per-face Jacobians and reconstructs the deformation via a Poisson solve, enabling smooth, globally coherent geometry updates and mitigating artifacts from noisy CLIP gradients (see Fig.~4 in the original paper):contentReference[oaicite:1]{index=1}. 
The method also employs a view-consistency loss to enforce multi-view coherence and a Jacobian regularization term to preserve source shape identity.
TextDeformer requires a pair of captions: a source caption describing the input mesh and an edited caption specifying the desired target geometry. 
To obtain these, we first construct the target caption following \cref{app:metrics}. 
We then generate a concise source caption from the original asset description using Gemini~2.5~Pro with the prompt below.
For each editing shape, we select the output from the \texttt{mesh\_best\_clip} directory as the final edited result. 
Since TextDeformer occasionally fails to optimize certain shapes (as noted in the original paper, where optimization may diverge or produce degenerate geometry), we exclude failure cases when computing the final averaged quantitative metrics.
\begin{tcolorbox}[colback=blue!5!white, colframe=blue!75!black, 
    title=Concise Source Caption, width=\textwidth, 
    boxrule=1pt, arc=2mm, auto outer arc,
    sharp corners=south,
    breakable,
    ]

    Write EXACTLY ONE short English noun phrase ($<$= 5 words) that summarizes the primary subject(s)
    
    described below. Use lower-case, no quotes, no punctuation.
    
    - examples:
    
      captions: 'a humanoid robot ...' -$>$ 'humanoid robot'
    
      captions: 'a low-poly house ...' -$>$ 'small house'

    captions: {captions\_text or ''},
    
    Now output the phrase:

\end{tcolorbox}

\item \noindent\textbf{Implementation Details of MVEdit.}
MVEdit~\cite{chen2024generic} is a training-free 3D editing framework that adapts pretrained 2D diffusion models into 3D-consistent multi-view editors using a 3D Adapter that jointly denoises multi-view images and reconstructs coherent geometry. 
For each editing task, the input consists of a source shape and an editing instruction, and the output is an edited 3D shape.
We reproduce MVEdit using the official Gradio-based implementation\footnote{\url{https://github.com/Lakonik/MVEdit}}. 
Specifically, we use the \texttt{Instruct 3D-to-3D} mode, which performs instruction-guided 3D shape editing.
We fix the random seed to 42 for all experiments, and keep all other hyperparameters identical to the ``polnareff'' example provided in the official interface. 
The edited shape produced by the pipeline is used as the final result.

\item \noindent\textbf{Implementation Details of EditP23.}
EditP23~\cite{bar2025editp23} is a mask-free and training-free 3D editing framework that propagates a user-provided 2D edit to a full multi-view editing.
We reproduce EditP23 following the official implementation\footnote{\url{https://github.com/editp23/editp23}}.
For each sample, we first render six multi-view images of the source shape using the rendering setup described in EditP23. 
Each editing run takes as input the source view, the edited target view, and the rendered multi-view grid. 
EditP23 outputs an edited multi-view grid, which we then reconstruct into a 3D mesh following the reconstruction pipeline detailed in the official implementation.
Following the configuration strategy in \cite{bar2025editp23}, we adopt different presets for different editing types. 
For style-change edits, we set the target guidance scale to 5.0 and use $n_{\text{max}} = 31$.
For all other editing categories, we set the target guidance scale to 21.0 and use $n_{\text{max}} = 39$.
The resulting reconstructed mesh is used as the final edited shape.

\item \noindent\textbf{Implementation Details of Direct Editing.}
Given the target image, we perform inference using the flow-based 3D generative model~\cite{hunyuan3d2025hunyuan3d} to obtain the edited shape.
All experiments follow the official implementation\footnote{\url{https://github.com/Tencent-Hunyuan/Hunyuan3D-2.1}}.

\item \noindent\textbf{Implementation Details of Editing-by-Inversion.}
Editing-by-Inversion extends Direct Editing~\cite{hunyuan3d2025hunyuan3d} by inserting a flow-based inversion prior to inference. 
We follow the conditional inversion procedure of UniEdit-Flow~\cite{jiao2025uniedit}. 
UniEdit-Flow introduces Uni-Inv, a predictor-corrector inversion method designed to more accurately recover the flow trajectory by mitigating the error accumulation inherent to straight, non-crossing rectified-flow paths.
In our setting, we perform conditional Uni-Inv using the source image as the conditioning input during inversion.
In specific, we set the target guidance scale to $1.0$ and run $50$ inversion steps to obtain an latent aligned with the source geometry, following the official implementation\footnote{\url{https://github.com/DSL-Lab/UniEdit-Flow}}.
After inversion, we perform standard flow-based inference using the editing image to synthesize the final edited shapes.

\item \noindent\textbf{Implementation Details of Inversion-free Editing.}
We reproduce the inversion-free editing~\cite{kulikov2024flowedit} on top of the flow-based 3D generative model~\cite{hunyuan3d2025hunyuan3d}.
It constructs a direct ODE between the source and target conditioned distributions, avoiding the Gaussian-noise traversal required in inversion-based editing. 
Concretely, it introduces a stochastic forward process that linearly interpolates between the source latent and random noise, and uses the resulting paired source-target latent to compute the velocity difference field, which is then averaged across multiple noise samples to update the editing trajectory. 
For fair comparison, we use the same hyperparameter settings as those in our method.
\end{itemize}

\section{Additional Experiments}\label{app:results}

\subsection{More Quantitative Results.}

\begin{table*}[t]
\centering
\caption{
\textbf{Quantitative Comparison across Different 3D Editing Methods on the Eval3DEdit benchmark.}
We report results using $\mathrm{Uni3D}_{\text{pc}} \uparrow$ for identity preservation and $\mathrm{Uni3D}_{\text{txt}} \uparrow$ for semantic modification, where higher values indicate better performance.
Metrics are evaluated across five representative editing categories, including action change, object addition, object removal, object replacement, and style change.
The Overall denotes the average performance across all categories.
}
\vspace{-1.0em}
\setlength{\tabcolsep}{2pt}
\small
\begin{tabular}{lcccccccccccc}
\toprule
\multirow{2}{*}{Method} & 
\multicolumn{2}{c}{Action Change} & 
\multicolumn{2}{c}{Object Addition} & 
\multicolumn{2}{c}{Object Removal} & 
\multicolumn{2}{c}{Object Replace.} & 
\multicolumn{2}{c}{Style Change} &
\multicolumn{2}{c}{Overall} \\
\cmidrule(lr){2-3} \cmidrule(lr){4-5} \cmidrule(lr){6-7} \cmidrule(lr){8-9} \cmidrule(lr){10-11} \cmidrule(lr){12-13}
& \scriptsize $\mathrm{Uni3D}_\text{pc}$ & \scriptsize $\mathrm{Uni3D}_\text{txt}$ 
& \scriptsize $\mathrm{Uni3D}_\text{pc}$ & \scriptsize $\mathrm{Uni3D}_\text{txt}$ 
& \scriptsize $\mathrm{Uni3D}_\text{pc}$ & \scriptsize $\mathrm{Uni3D}_\text{txt}$ 
& \scriptsize $\mathrm{Uni3D}_\text{pc}$ & \scriptsize $\mathrm{Uni3D}_\text{txt}$ 
& \scriptsize $\mathrm{Uni3D}_\text{pc}$ & \scriptsize $\mathrm{Uni3D}_\text{txt}$ 
& \scriptsize $\mathrm{Uni3D}_\text{pc}$ & \scriptsize $\mathrm{Uni3D}_\text{txt}$ \\
\midrule
\multicolumn{3}{l}{\textcolor{gray}{\textit{Optimization-based 3D Editing Methods}}} & \multicolumn{10}{l}{} \\
TextDeformer~\cite{gao2023textdeformer} & 0.1321 & 0.1393 & 0.2170 & 0.1293 & 0.1549 & 0.1249 & 0.2579 & 0.1269 & 0.1444 & 0.1288 & 0.1818 & 0.1298 \\
\midrule
\multicolumn{3}{l}{\textcolor{gray}{\textit{LRM-based 3D Editing Methods}}} & \multicolumn{10}{l}{} \\
MVEdit~\cite{chen2024generic} & 0.4891 & 0.1204 & 0.1366 & 0.0653 & 0.2224 & 0.0571 & 0.2741 & 0.0859 & 0.2162 & 0.0685 & 0.2677 & 0.0794 \\
EditP23~\cite{bar2025editp23} & 0.1272 & 0.1527 & 0.0879 & 0.0667 & 0.0733 & 0.0774 & 0.1240 & 0.0862 & 0.0887 & 0.0743 & 0.1002 & 0.0915 \\
\midrule
\multicolumn{3}{l}{\textcolor{gray}{\textit{LFM-based 3D Editing Methods}}} & \multicolumn{10}{l}{} \\
Direct Editing~\cite{hunyuan3d2025hunyuan3d} & 0.2611 & 0.2227 & 0.2982 & 0.1878 & 0.2658 & 0.1184 & 0.4006 & 0.2250 & 0.2485 & 0.1274 & 0.2948 & 0.1762 \\
Editing-by-Inversion~\cite{jiao2025uniedit} & 0.4512 & 0.2432 & 0.4990 & 0.2445 & 0.5094 & 0.2320 & 0.5807 & 0.2767 & 0.5317 & 0.2609 & 0.5144 & 0.2515 \\
Inversion-free Editing~\cite{kulikov2024flowedit} & 0.4701 & 0.2375 & 0.5242 & 0.2377 & 0.5447 & 0.2436 & 0.6052 & 0.2742 & 0.5407 & 0.2566 & 0.5370 & 0.2499 \\
\noalign{\vskip 0.25em}
\hdashline
\noalign{\vskip 0.25em}
AnchorFlow (Ours) & 0.3123 & 0.2756 & 0.4438 & 0.2504 & 0.4501 & 0.2314 & 0.5220 & 0.2788 & 0.4577 & 0.2494 & 0.4372 & 0.2571 \\
\bottomrule
\end{tabular}
\label{tab:comp-uni3d}
\end{table*}

% \multicolumn{3}{l}{\textcolor{gray}{\textit{Optimization-based 3D Editing Methods}}} & \multicolumn{10}{l}{} \\
% TextDeformer~\cite{gao2023textdeformer} & 0.1321 & 0.1393 & 0.2170 & 0.1293 & 0.1549 & 0.1249 & 0.2579 & 0.1269 & 0.1444 & 0.1288 & 0.1818 & 0.1298 \\
% \midrule
\noindent\textbf{Quantitative Evaluation with Uni3D-based Metrics.}
Beyond the CLIP-based evaluation in main paper, we further validate these observations using Uni3D-based identity and semantic metrics (\cf \cref{app:metrics-uni3d}), as reported in \cref{tab:comp-uni3d}. 
The results show a consistent trend in \cref{tab:comp-all}.
LRM-based methods exhibit the weakest performance across both identity preservation and semantic modification due to multi-view inconsistency. 
LFM-based approaches achieve substantially higher scores, confirming the benefit of operating directly in a 3D latent space. 
Within this category, editing-by-inversion and inversion-free editing obtain strong identity preservation but either rely on expensive inversion or tend to produce under-edited or distorted geometry. 
In contrast, our AnchorFlow method achieves competitive identity scores while delivering the strongest semantic alignment among all LFM methods, showing that our approach provides a more balanced and reliable editing results. 
These Uni3D results corroborate the CLIP-based findings and further demonstrate the effectiveness and robustness of our training-free 3D editing pipeline.

\begin{figure}[t]
    \centering
    \includegraphics[width=1.0\linewidth]{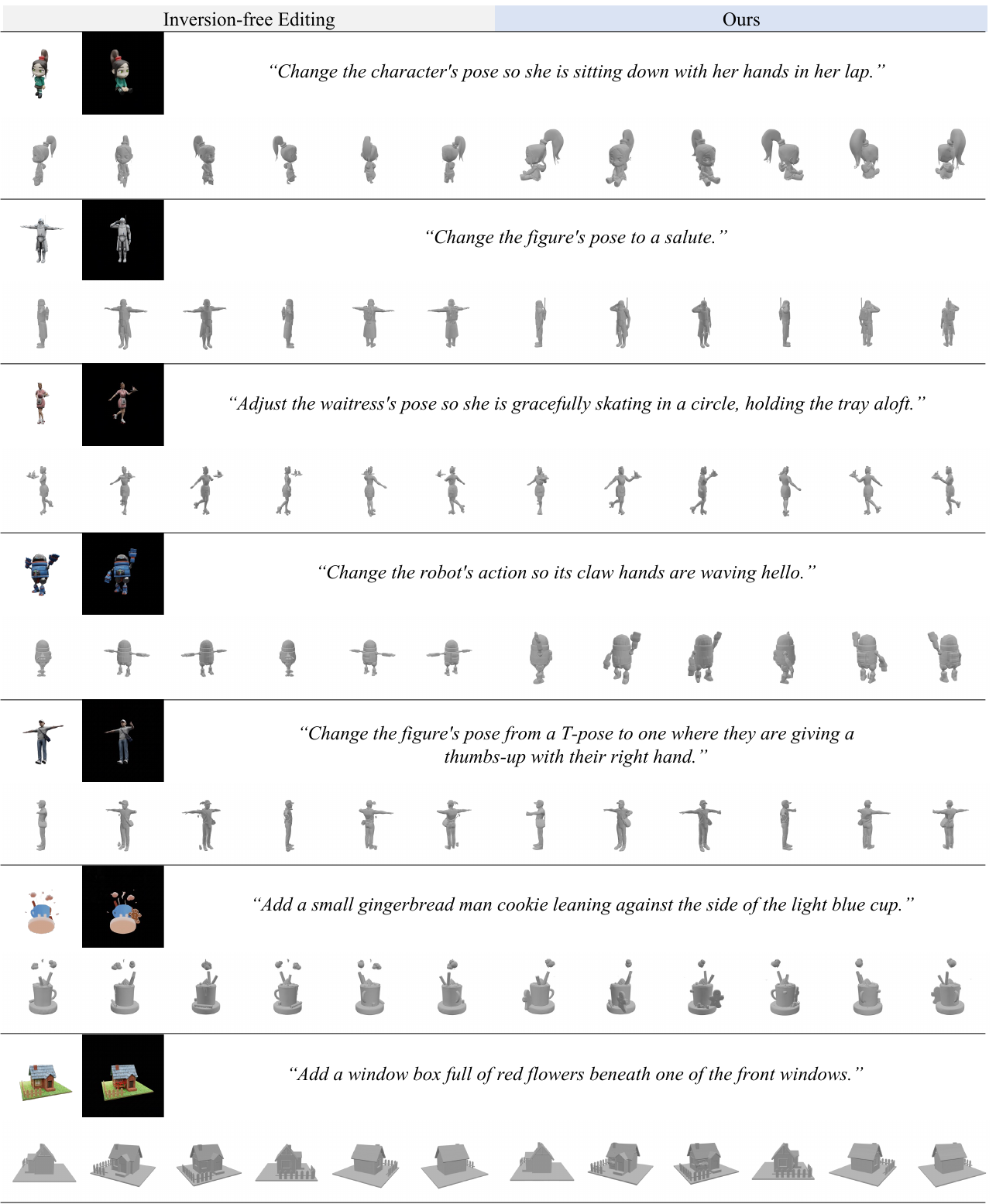}
    \caption{More qualitative results using AnchorFlow.}
    \label{fig:app_comp_1}
\end{figure}

\begin{figure}[t]
    \centering
    \includegraphics[width=1.0\linewidth]{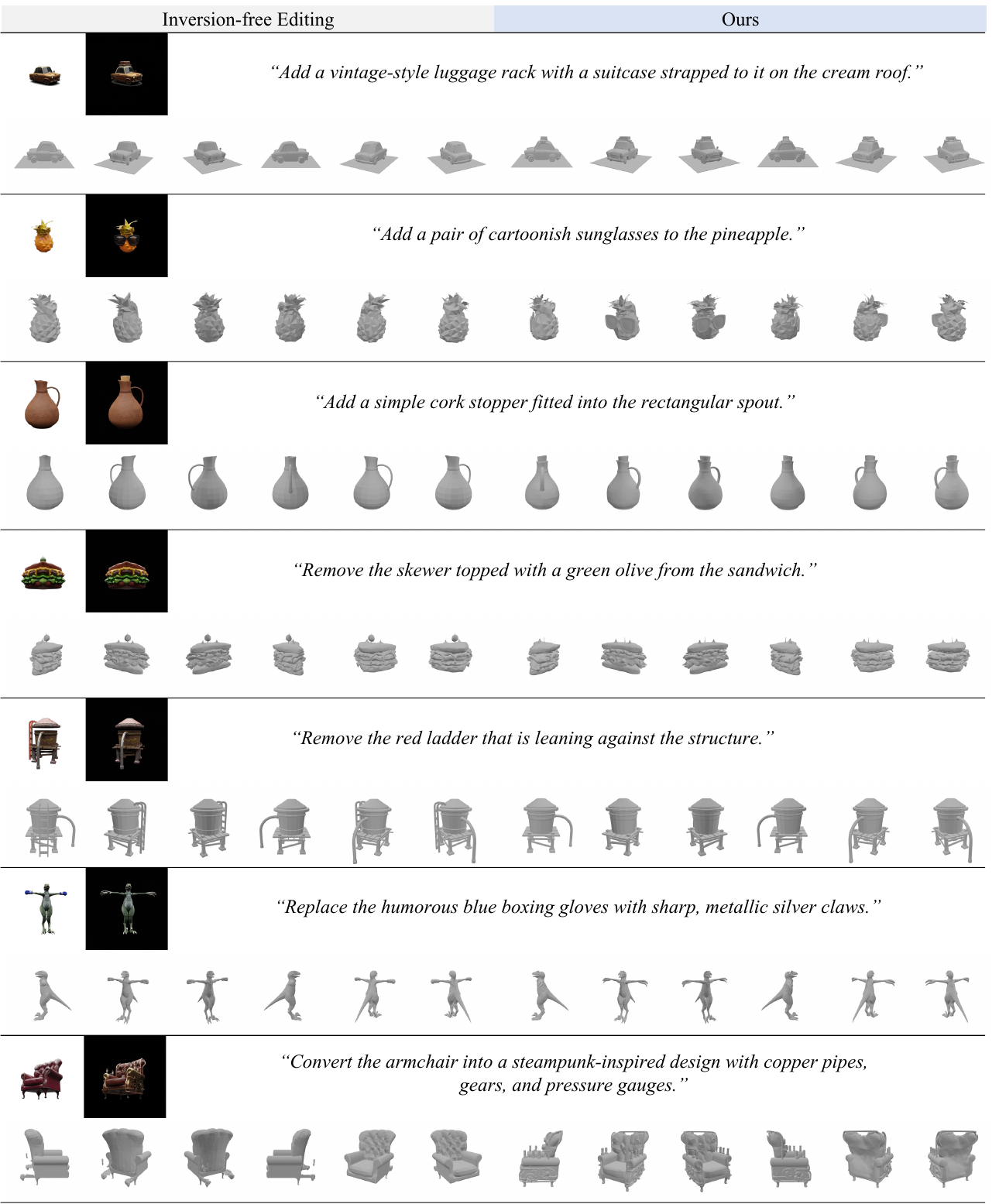}
    \caption{More qualitative results using AnchorFlow.}
    \label{fig:app_comp_2}
\end{figure}

\subsection{More Qualitative Results.}
Beyond the qualitative comparisons provided in the main paper, we include additional examples in \cref{fig:app_comp_1} and \cref{fig:app_comp_2} to further assess the effectiveness of our method. 
Note that in all comparative experiments, AnchorFlow and the baseline~\cite{kulikov2024flowedit} share same hyperparameter settings to ensure a fair comparison.
The results demonstrate that our method produces edits that are both semantically faithful to the instruction and geometrically coherent. 
Across non-rigid editing tasks such as action change (\eg, rows 1--5 in \cref{fig:app_comp_1}), AnchorFlow generates smooth and globally consistent articulations. 
In contrast, inversion-free editing often exhibits insufficient deformation (\eg, rows 2 and 4 in \cref{fig:app_comp_1}) or noticeable distortions in limb shape or body proportions (\eg, rows 1 and 5 in \cref{fig:app_comp_1}). 
These additional examples confirm that our method maintains stable geometry even under large pose changes while accurately following fine-grained instructions such as ``saluting'' and ``waving.''
For rigid editing tasks, such as object addition (\eg, rows 1--3 in \cref{fig:app_comp_2}), AnchorFlow performs precise and localized modifications without affecting unrelated regions. 
This further supports that AnchorFlow enables mask-free editing.
Overall, the extended qualitative results provide additional evidence that AnchorFlow achieves strong semantic modification while preserving identity across both rigid and non-rigid editing tasks.

\end{document}